\documentclass[journal]{IEEEtran}
\usepackage{amsmath,amsfonts}

\usepackage{algorithmic}
\usepackage{algorithm}
\usepackage{array}
\usepackage[caption=false,font=normalsize,labelfont=footnotesize,textfont=sf]{subfig}
\usepackage{textcomp}
\usepackage{stfloats}
\usepackage{url}
\usepackage{verbatim}
\usepackage{graphicx}
\usepackage{cite}
\usepackage{orcidlink}
\usepackage{bm}
\usepackage{amssymb}
\usepackage{xcolor}
\usepackage{paralist}
\usepackage{tabularx}
\usepackage{mathtools}

\newcommand{\etal}{{et~al.}}

\newcommand{\ie}{{\it i.e.}}

\newcommand{\reals}{\mathbf{R}}

\newcommand{\calD}{\mathcal{D}}

\newtheorem{definition}{Definition}

\newtheorem{proposition}{Proposition}
\newtheorem{remark}{Remark}

\begin{document}

\title{Reinforcement Learning of CPG-regulated Locomotion Controller for a Soft Snake Robot}

\author{Xuan Liu \orcidlink{0000-0003-4373-1359}, \IEEEmembership{Student Member, IEEE}, Cagdas D. Onal \orcidlink{0000-0002-3307-1273}, \IEEEmembership{Member, IEEE}, and Jie Fu \orcidlink{0000-0002-4470-2827}, \IEEEmembership{Member, IEEE} 
        
\thanks{This work was supported in part by the National Science Foundation under grant \#1728412. \textit{(Corresponding author: Jie Fu)} }
\thanks{Xuan Liu and Cagdas Onal are with the Robotics Engineering Department at Worcester Polytechnic Institute, Worcester, MA, US (e-mail: xliu9@wpi.edu; cdonal@wpi.edu).}
\thanks{Jie Fu is with the Department of Electrical and Computer Engineering, University of Florida, Gainesville, FL, US (e-mail: fujie@ufl.edu, fax number: 352-392-8671).}
}





\maketitle

\begin{abstract}
Intelligent control of soft robots is challenging due to the nonlinear and difficult-to-model dynamics. One promising model-free approach for soft robot control is reinforcement learning (RL). However, model-free RL methods tend to be computationally expensive and data-inefficient and may not yield natural and smooth locomotion patterns for soft robots. In this work, we develop a bio-inspired design of a learning-based goal-tracking controller for a soft snake robot. The controller is composed of two modules: An RL module for learning goal-tracking behaviors given the unmodeled and stochastic dynamics of the robot, and a central pattern generator (CPG) with the Matsuoka oscillators for generating stable and diverse locomotion patterns. We theoretically investigate the maneuverability of Matsuoka CPG's oscillation bias, frequency, and amplitude for steering control, velocity control, and sim-to-real adaptation of the soft snake robot. Based on this analysis, we proposed a composition of RL and CPG modules such that the RL module regulates the tonic inputs to the CPG system given state feedback from the robot, and the output of the CPG module is then transformed into pressure inputs to pneumatic actuators of the soft snake robot. This design allows the RL agent to naturally learn to entrain the desired locomotion patterns determined by the CPG maneuverability. We validated the optimality and robustness of the control design in both simulation and real experiments, and performed extensive comparisons with state-of-art RL methods to demonstrate the benefit of our bio-inspired control design.
\end{abstract}


\section{Introduction}
\IEEEPARstart{D}{ue} to their flexible geometric shapes and deformable materials, soft continuum robots have great potential to perform tasks in dangerous and cluttered environments, including natural disaster relief and pipe inspection \cite{majidi2014soft}. 
However, planning and control of such robots are  challenging, as these robots have infinitely many degrees of freedom in their body links, and soft actuators with stochastic, unknown dynamics and delayed responses.  

\IEEEpubidadjcol
In this work, we develop a bio-inspired locomotion controller for soft robot snakes to achieve serpentine-like locomotion for set-point tracking tasks. Specifically, we consider utilizing the properties of CPGs, which consists of a special group of neural circuits that are able to generate rhythmic and non-rhythmic activities for organ contractions and body movements in animals. Such activities can be activated, modulated, and reset by neuronal signals mainly from two directions: bottom-up ascendant feedback information from afferent sensory neurons or top-down descendant signals from high-level modules including mesencephalic locomotor region (MLR) \cite{ijspeert2008central} and motor cortex \cite{roberts1998central, yuste2005cortex}.  

\begin{figure}[h!]
    \centering
    \includegraphics[width=0.9\columnwidth]{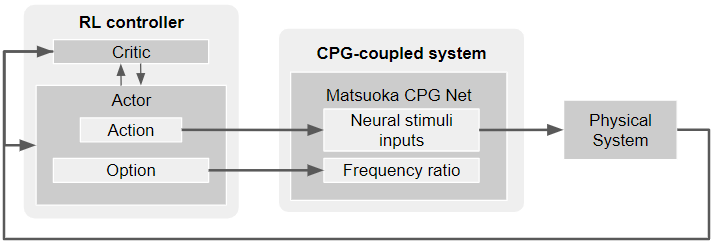}
    \caption{Schematic view of learning-based CPG controller.}
    \label{fig:schematic_view}
 \end{figure}

In literature, bio-inspired control methods have been studied for the control design of rigid robots' locomotion, including legged\cite{mori2004reinforcement, endo2008learning, nassour2014multi, dzeladini2018cpg, 8585406} and serpentine locomotion \cite{crespi2005swimming, crespi2008online, ryu2010locomotion,  bing2017towards, wang2017cpg, bing2019end}. The general approach is to generate motion patterns mimicking animals' behaviors and then track these trajectories with a closed-loop control design. In \cite{ijspeert2008central}, the authors developed a trajectory generator for a rigid salamander robot using Kuramoto CPGs and used low-level PD controllers to track the desired motion trajectories generated by the oscillator. Ryu \etal \cite{ryu2010locomotion} established the velocity control CPG by adapting its frequency parameter with additional linear dynamics. In \cite{wang2017cpg}, the authors introduced a control loop that adjusts the oscillation patterns including frequency, amplitude, and phase of the oscillation for adapting to the changes in the terrain. Their results show the advantage of the Hopf oscillator on the direct access to the oscillation patterns for different locomotion purposes. In \cite{6420917}, the Matsuoka oscillator is combined with the amplitude modulation method to realize steering control of a rigid snake robot. However, these approaches have not provided a way to maneuver the oscillation patterns intelligently. Recent work has combined learning-based method and CPG systems for control of rigid robotic systems. Sartoretti \etal \cite{8755279} proposed a decentralized approach, where each actuator of an articulated rigid snake robot is controlled independently by a neural network (NN) controller learned with an end-to-end RL algorithm. Another recent work \cite{bing2019end} employed a spiking neural net (SNN) under the regulation of reward-modulated spike-timing-dependent plasticity (R-STDP) to map visual information into wave parameters of a phase-amplitude CPG net, which generates desired oscillating patterns to locomote a rigid snake robot chasing a red ball. The similar idea of combining learning and CPG is also investigated in the legged robot. Tran \etal \cite{8585406} employed a Q-learning selector to make decision on switching among different CPG patterns in a disturbance recovery task during bipedal locomotion. In \cite{9932888}, the authors proposed a CPG-RL method that directly learns the neural oscillator's intrinsic amplitude and frequency, and coordinate the decoupled oscillator network to control the legged locomotion of a quadruped robot.

For a rigid snake robot, existing literature \cite{ijspeert2008central} has introduced a model-based control design combined with CPG for motion planning. Despite the success of bio-inspired control with rigid snake robots, the same control scheme may not work as desired for soft snake robots. This is because, in these approaches, the trajectories generated by CPG require high-performance low-level controllers for tracking. The tracking performance cannot be reproduced with soft snake robots due to the nonlinear, delayed, and stochastic dynamical response from the soft actuators.

To this end, we develop a bio-inspired learning-based control framework for soft snake robots with two key components: 
To achieve intelligent and robust goal-tracking with changing goals, we use model-free RL \cite{sutton1999policy,schulman2017proximal} to map the feedback of soft actuators and the goal location, into control commands of a CPG network. The CPG network consists of coupled Matsuoka oscillators \cite{matsuoka1985sustained}. The Matsuoka CPG network acts as a low-level motion controller to generate actuation inputs directly to the soft snake robots for achieving smooth and diverse motion patterns. The two networks form a variant of cascade control with only one outer-loop, as illustrated in Fig.~\ref{fig:schematic_view}. Comparing to other neural oscillators used in \cite{8755279, bing2019end, 9932888}, the Matsuoka oscillator has the following special properties
\begin{enumerate}
    \item It is in the class of half-center\cite{brown1911} oscillator model that describes mutually inhibiting mechanism in a pair of neurons. Such mechanism produces alternate activities of flexors and extensors, which can be used to directly control a pair of actuators mimicking antagonistic muscles;
    \item It has clear boundary conditions for the parameters such that the neurons can generate free-response oscillation when satisfying the boundary condition\cite{matsuoka1987mechanisms}; 
    \item On the basis of free-response oscillation, the entrainment property \cite{matsuoka2011analysis, Matsuoka2013FrequencyRO} allows the intelligent controller to autonomously regulate the oscillation pattern of the system with forced-response oscillation input;
    \item It is a piece-wise linear system with local linearity in certain quadrants.
\end{enumerate} 

Based on the above properties of the Matsuoka oscillator, we showed that several dynamic properties of the Matsuoka oscillators can be leveraged in designing the interconnection between RL and CPG. We have proved that the \emph{steering control} can be realized by modulating both the amplitudes bias and duty cycles of the neural stimuli inputs of the CPG network, and the \emph{velocity control} can be realized by tuning the oscillating frequencies of the CPG net. These findings enable us to flexibly control the slithering locomotion with a CPG network given state feedback from the soft snake robot and the control objective.

This paper is an extension of our preliminary work\cite{xliu2020} that designs a learning-based set-point tracking control for soft snake robots. In comparison to \cite{xliu2020}, we make the following improvements:

\begin{enumerate}
    \item \textbf{Theoretical analysis of steering maneuverability:} We analyze the property of the biased oscillation in the Matsuoka oscillator. Using describing function analysis, we show that when the tonic inputs of the Matsuoka oscillator are bounded and satisfy certain constraints, the bias of the output signal becomes linearly related to the tonic inputs. This feature makes the steering control of the snake robot easier to learn for an RL agent.

    \item \textbf{Free-response Oscillation Constraints (FOC) for sim-to-real transfer:} We investigate the transient property of the Matsuoka Oscillator from free-response oscillation to forced-response oscillation. Using this property, we introduce a fixed free-response tonic input signal to help regulate the amplitude and oscillation frequency of the forced tonic inputs which are generated by the RL policy. The new approach is referred to as Free-response Oscillation Constrained Proximal Policy Optimization Option-Critics with Central Pattern Generator (FOC-PPOC-CPG). This approach improves the transferability of the RL control policy learned in the simulation to the real robot.

    \item \textbf{Improve reward density with potential field function:} We newly introduce a potential-field-based reward shaping to accelerate the learning process.
    \item \textbf{Comprehensive sim-to-real tests and analysis: } We added new experiments comparing the learning efficiency and adaptability of the policy between the proposed method and vanilla Proximal Policy Optimization (PPO) \cite{schulman2017proximal}. Based on the experimental results for both simulation and reality, we show that our soft snake robot equipped with a properly designed “vertebrate” (the CPG system) can be more easily controlled by the RL agent. Our approach also achieves more reliable locomotion performance under various goal-reaching locomotion tasks that are unseen during the training process. 

\end{enumerate}

The paper is structured as follows: 
Section~\ref{sec:sys} provides an overview of the robotic system and the state space representation. Section~\ref{sec:CPG} presents the design and configuration of the CPG network. Section~\ref{sec:dis} discusses the key properties of the CPG network and the relation to the design of an artificial neural network for the RL module. Section~\ref{sec:sol} introduces a curriculum and reward design for learning a goal-tracking locomotion controller with a soft snake robot. Section~\ref{sec:result} presents the experimental validation and evaluation of the controller in both simulated and real snake robots.

\section{System Overview of the Soft Snake Robot}
\label{sec:sys}

\begin{figure}[ht]
    \centering
    \includegraphics[width=0.9\columnwidth]{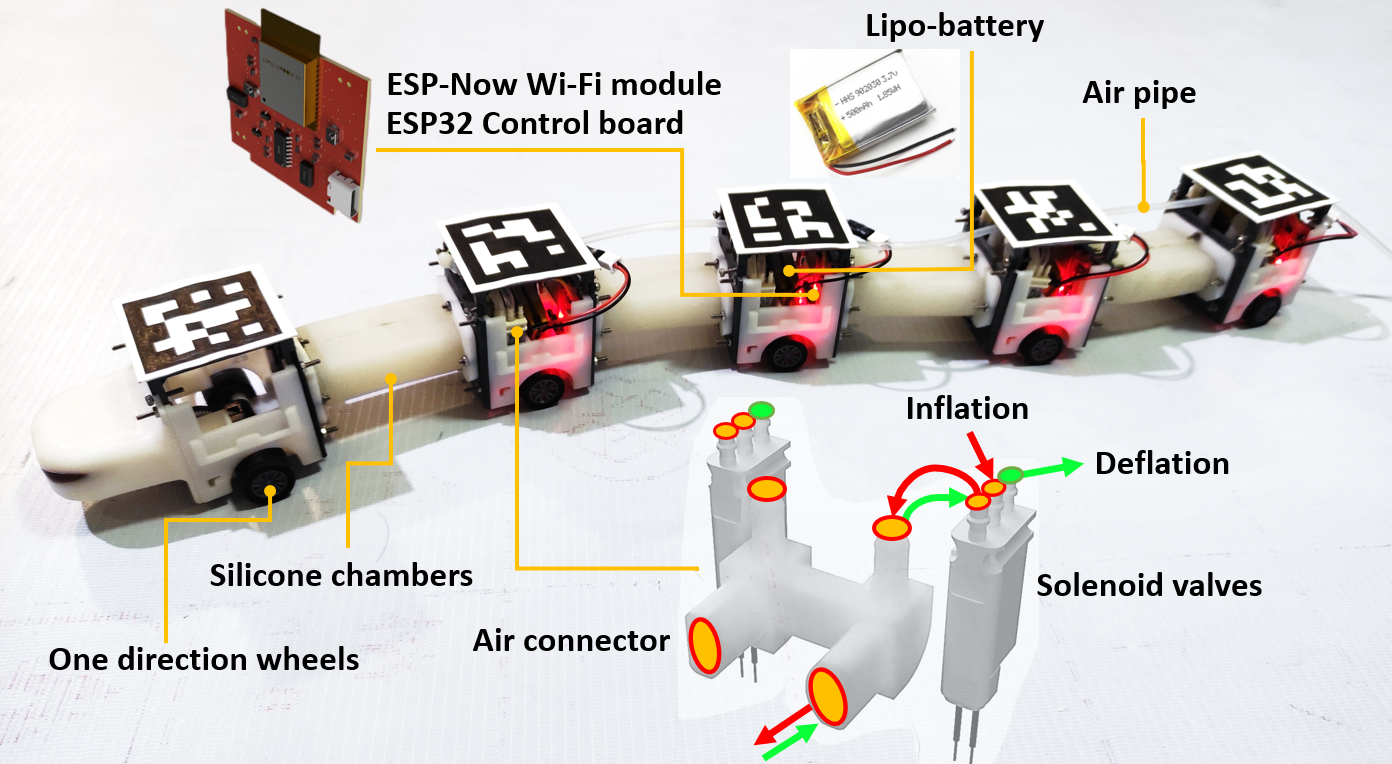}
    \caption{Mechatronics design of the soft snake robot.}
    \label{fig:mechatronics}
\end{figure}

\begin{figure*}[ht!]
    \centering
    \includegraphics[width=0.9\textwidth]{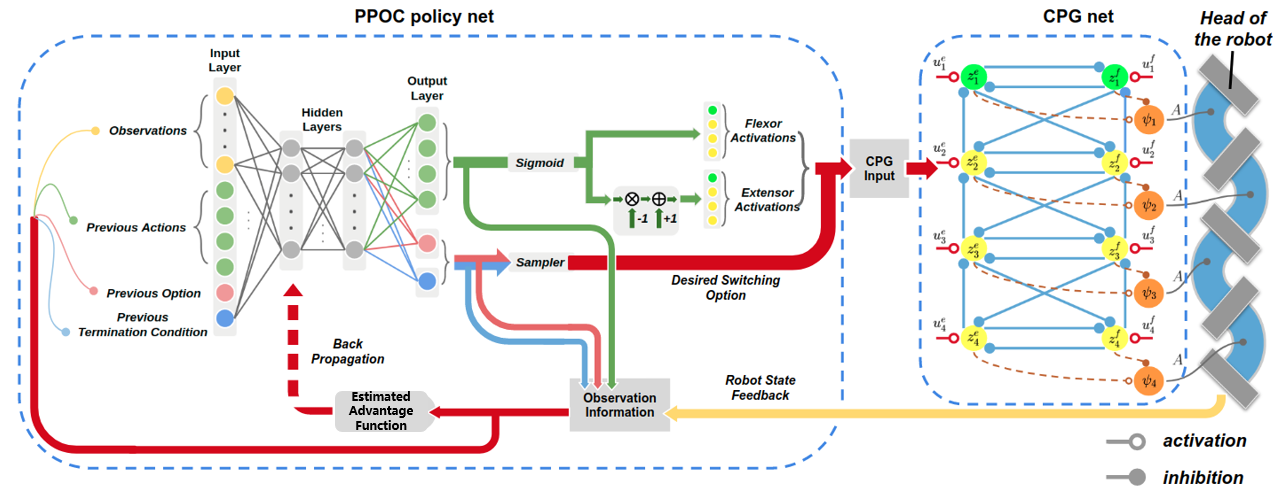}
    \caption{Illustrating the input-output connection of the PPOC-CPG net.}
    \label{fig:rlmatsuoka}
\end{figure*}

As shown in Fig.~\ref{fig:mechatronics}, our soft snake robot is a subtype of WPI-SRS series robot\cite{mluo2020}. It consists of $4$ pneumatically actuated soft links. The soft links are made of Ecoflex\texttrademark~00-30 silicone rubber. Each soft link of the robot has two air chambers mimicking antagonistic muscle (detailed structure of the soft body can be found in \cite{luo2017toward, mluo2020}). The links are connected through rigid bodies enclosing the electronic components that are necessary to control the snake robot. Each rigid body contains an ESP32 module (powered by a Lithium-polymer battery) for control command communication and a pair of SMC-S070C-SCG solenoid valves that control the inflation and deflation of the air chambers. Only one chamber on each link is active (pressurized) at a time. In addition, the rigid body components have a pair of one-direction wheels to model the anisotropic friction of real snakes.

\begin{figure}[ht]
    \centering
    \includegraphics[width=0.85\columnwidth]{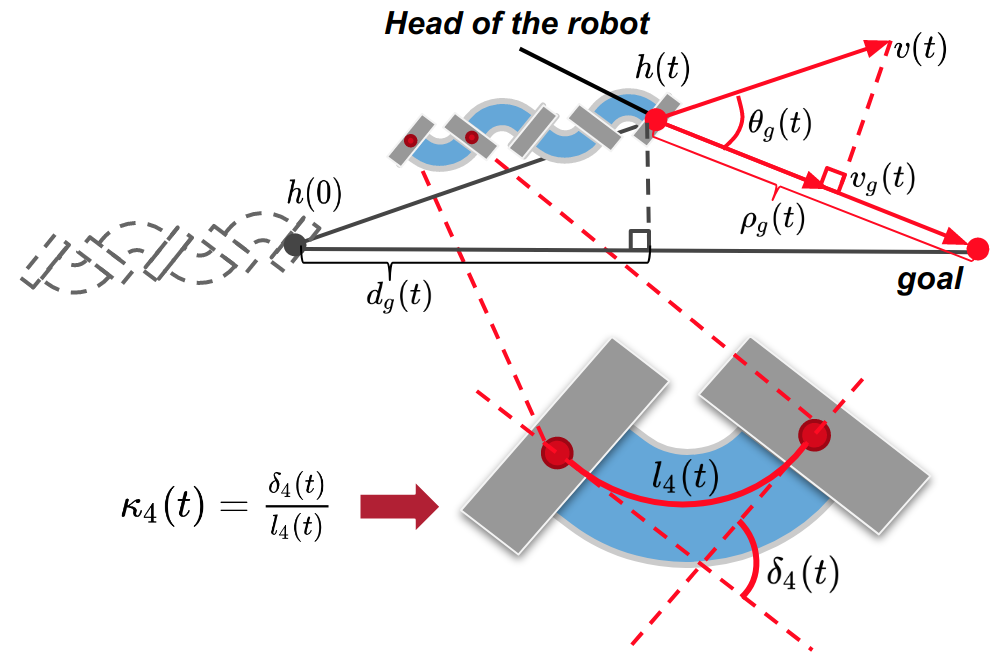}
    \caption{Notation of the state space configuration of the robot.}
    \label{fig:coordinate}
\end{figure}

The configuration of the robot's coordinate is shown in Figure~\ref{fig:coordinate}. At time $t$, state $\bm{h}(t) \in \reals^2$ is the planar Cartesian position of the center of mass (COM) of the snake's head, $\bm{\rho_g}(t) \in \reals^2$ is the planar displacement vector pointing from snake's head COM to the goal position, $d_g(t)\in \reals$ is the distance traveled along the head-to-goal-direction from the initial head COM position, $\bm{v}(t) \in \reals^2$ is the instantaneous planar velocity vector of the snake's head COM, $\theta_g(t)$ is the angle between vector $\bm{\rho_g}(t)$ and vector $\bm{v}(t)$, and the locomotion speed $v_g(t) \in \reals$ is the length of the projection of $\bm{v}(t)$ on the head-to-goal-direction. According to \cite{luo2014theoretical}, the bending curvature of each body link at time $t$ is computed by
$
    \kappa_i(t) = \frac{\delta_i(t)}{l_i(t)}, \text{ for } i=1,\ldots, 4,
$
where $\delta_i(t)$ and $l_i(t)$ are the relative bending angle and the length of the middle line of the $i$-th soft body link.

In \cite{renato2019}, we developed a physics-based simulator that models the inflation and deflation of the air chamber and the resulting deformation of the soft bodies with tetrahedral finite elements. The simulator runs in real time using GPU. We use the simulator for learning the locomotion controller in the soft snake robot, and then apply the learned controller to the real robot.

\begin{figure*}[ht!]
    \centering
    \includegraphics[width=0.66\textwidth]{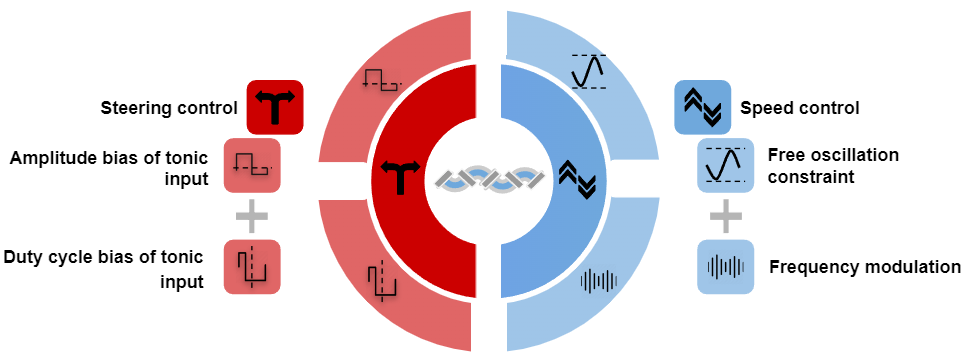}
    \caption{An overview of the maneuverability of Serpentine locomotion with the Matsuoka oscillator.}
    \label{fig:steerwithmatsuoka}
\end{figure*}

\section{Design of a CPG Network for the Soft Snake Robot Locomotion}
\label{sec:CPG}

In this section, we introduce our CPG network design consisting of interconnected Matsuoka oscillators \cite{matsuoka1987mechanisms, matsuoka2011analysis}. 

\noindent \textbf{Primitive Matsuoka CPG: }A primitive Matsuoka CPG consists of a pair of mutually inhibited neuron models. The dynamical model of the primitive Matsuoka CPG is given as follows:
\begin{align}
\begin{split}
    &K_f \tau_r \Dot{x}_i^e = -x_i^e - a z_i^f - b y_i^e - \sum_{j=1}^N w_{ji}y_j^e + u_i^e + c,\\ 
    &K_f \tau_a \Dot{y}_i^e = z_i^e - y_i^e,\\ 
    &K_f \tau_r \Dot{x}_i^f = -x_i^f - a z_i^e - b y_i^f - \sum_{j=1}^N w_{ji}y_j^f + u_i^f + c,\\ 
    &K_f \tau_a \Dot{y}_i^f = z_i^f - y_i^f,
\end{split}\label{eq:matsuoka}
\end{align} 
where the subscripts $e$ and $f$  represent variables related to the extensor neuron and flexor neuron, respectively. The tuple $(x_i^q, y_i^q)$, $q \in  \{e,f\}$ represents the activation state and self-inhibitory state of $i$-th neuron respectively,  $z_i^q = g(x_i^q) = \max(0, x_i^q)$\footnote{The maximum function is noted as $g(\cdot)=\max(0, \cdot)$ in this paper.} is the output of $i$-th neuron,  $b \in \reals$ is a weight parameter, $u_i^e, u_i^f$ are the forced tonic inputs to the oscillator, and  $K_f \in \reals$ is the frequency ratio.
The set of parameters in the system includes the discharge rate $\tau_r \in \reals$, the adaptation rate $\tau_a \in \reals$,  the mutual inhibition weights between flexor and extensor $a\in \reals$, the inhibition weight $w_{ji}\in \reals$ representing the coupling strength with the neighboring primitive oscillator, and the free-response oscillation tonic input $c\in \reals$ ($c=0$ in our previous work\cite{xliu2020}). In our system, all coupled signals including $x_i^q, y_i^q$ and $z_i^q$ ($q \in  \{e,f\}$) are inhibiting signals (negatively weighted), and only the tonic inputs are activating signals (positively weighted). In the current system, we have $N=4$ primitive Matsuoka CPGs. 
For simplicity, we introduce a vector \begin{equation}
\label{eq:bvec}
    \bm{u} = [u_1^e, u_1^f, u_2^e, u_2^f, u_3^e, u_3^f, u_4^e, u_4^f]^T
\end{equation} to represent all tonic inputs to the CPG net.

\noindent \textbf{Structure of the Matsuoka CPG Network for the Soft Snake Robot: } Extending from a primitive Matsuoka CPG system to the multi-linked snake robot, we construct a CPG network shown on the right of Fig.~\ref{fig:rlmatsuoka}. The network includes four linearly coupled primitive Matsuoka oscillators. 
It is an inverted, double-sized version of Network VIII introduced in Matsuoka's paper \cite{matsuoka1987mechanisms}. The network includes four pairs of nodes. Each pair of nodes (e.g., the two nodes colored green/yellow) in a row represents a primitive Matsuoka CPG \eqref{eq:matsuoka}. The edges correspond to the coupling relations among the nodes. In this graph, all the edges with hollowed endpoints are positive activating signals, while the others with solid endpoints are negative inhibiting signals. The oscillators are numbered $1 \text{ to } 4$ from head to tail of the robot. In order to build the connection between the CPG network and robot actuators, we define the output of the $i$-th primitive Matsuoka CPG as
\begin{align}
    \psi_i = a_{\psi} z_i = a_{\psi} (z_i^e - z_i^f),
\end{align}
where $a_{\psi}$ is a ratio coefficient of $z_i$. Given the Bounded Input Bounded Output (BIBO) stability of the Matsuoka CPG net \cite{matsuoka1985sustained}, the outputs $\bm{\psi} = [\psi_1,\psi_2,\psi_3,\psi_4]^T$ from the primitive oscillators can be limited within $[-1, 1]$ by adjusting the ratio $a_{\psi}$. We let $\psi_i = 1$ for the full inflation of the $i$-th extensor actuator and zero inflation of the $i$-th flexor actuator, and vice versa for $\psi_i = -1$. The actual pressure input to the $i$-th chamber is $\lambda_i\cdot\psi_i$, where $\lambda_i$ is the maximal pressure input of each actuator. The primitive oscillator with green nodes controls the oscillation of the head joint. This head oscillator also contributes as a rhythm initiator in the oscillating system, followed by the rest parts oscillating with different phase delays in sequence. Figure~\ref{fig:rlmatsuoka} shows all activating signals to the CPG network.

\noindent \textbf{Configuring the Matsuoka CPG Network: } To determine the hyper-parameters in the CPG network that generate a more efficient locomotion pattern, we employed a genetic programming (GP) algorithm similar to \cite{ijspeert1999evolving}. In this step, all tonic inputs are assigned with value $1$ for the simplicity of fitness evaluation. 

We define the fitness function--the optimization criteria--in GP as $F(v_{g, T}, \theta_{g, T}, d_{g, T}) = a_1 |v_{g, T}| - a_2 |\theta_{g, T}| + a_3 |d_{g, T}|, $ where $g$ indicates a fixed goal initiated in the heading direction of the snake robot, $T$ indicates the terminating time of fitness evaluation for each trial, and all coefficients $a_1, a_2, a_3 \in \reals^+$ are constants\footnote{In experiments, the following  parameters are used: $a_1= 40.0$, $a_2= 100.0$, $a_3= 50.0$, and $T=6.4 \text{ sec}$. }.

To achieve stable and synchronized oscillations of the whole system, the following constraint must be satisfied \cite{matsuoka1985sustained}:
\begin{align}
\label{eq:oscilexist}
    (\tau_a-\tau_r)^2 < 4\tau_r\tau_a b,
\end{align}
where $\tau_a, \tau_r, b > 0$. To satisfy this constraint, we can set the value of $b$ much greater than both $\tau_r$ and $\tau_a$, or make the absolute difference $|\tau_r-\tau_a|$ sufficiently small.

In other words, this fitness function is a weighted sum over the robot's instantaneous speed, deviation angle, and total traveled distance on a fixed straight line at the terminating time $T$. In this scenario, a better-fitted configuration is supposed to maintain oscillating locomotion and reaches faster locomotion speed $|v_{g, T}|$ along the original heading direction at time $T$. In addition, the locomotion pattern is required to have a smaller deviation between the robot's heading direction and the goal direction (with a small $ |\theta_g|$), and with overall a longer traveled distance along the robot's heading direction ($|d_g|$).

The desired parameter configuration found by GP is given in Table.~\ref{tab:config} in Appendix~\ref{sec:data}.

\section{Maneuverability Analysis and Design of the Learning-based Controller with the Matsuoka CPG Network}
\label{sec:dis}

When provided with equally constant tonic inputs, the designed Matsuoka CPG system can generate stable oscillation patterns to efficiently drive the soft snake robot slithering forward. However, a single CPG network cannot achieve intelligent locomotion and goal-tracking behaviors with potentially time-varying goals. For an intelligent controller, the free turning and accelerating (or decelerating) behaviors are the fundamental skills to realize autonomous locomotion in the goal-tracking tasks. In this paper, we denote these two maneuverability demands as -- steering control and velocity control 
(see Fig.~\ref{fig:steerwithmatsuoka}). The later parts will focus on investigating the properties of the Matsuoka CPG system to prove that it is controllable from both steering and velocity control perspectives. We design a proper connection between RL actions and controllable coefficients of the Matsuoka CPG system so that both steering and velocity control of the snake robot can be efficiently learned by the RL agent. 

For steering control, we prove that the bias of tonic inputs is linearly proportional to the bias of the CPG output in both amplitude and duty cycle dimensions. This property inspires a rule that transforms the action outputs of the RL policy into the tonic inputs of the CPG system.

Next, we excavate two mechanisms that are helpful for velocity control. First, we show that the frequency ratio coefficient $K_f$ allows the RL agent to tune the locomotion velocity by directly adjusting the oscillation frequency. Second, by introducing the free-response oscillation constraint, we provide a way to adjust the converging amplitude of the oscillation driven by the RL agent. With experiments, we show that the free-response oscillation constraint is very helpful for reducing performance drop in the sim-to-real problem.

\begin{figure}[htbp]
\centering
\subfloat[]{\includegraphics[width=0.79\columnwidth]{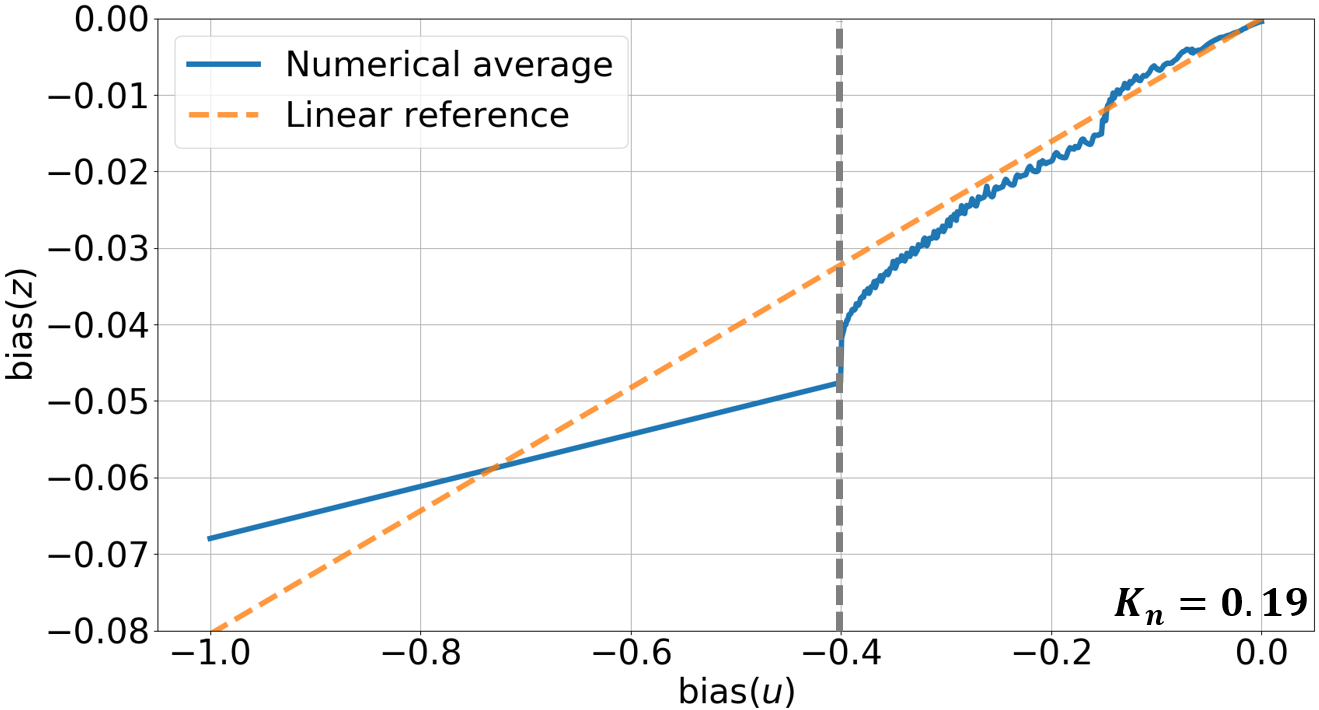}%
\label{fig:bias19}}\\ 
\subfloat[]{\includegraphics[width=0.79\columnwidth]{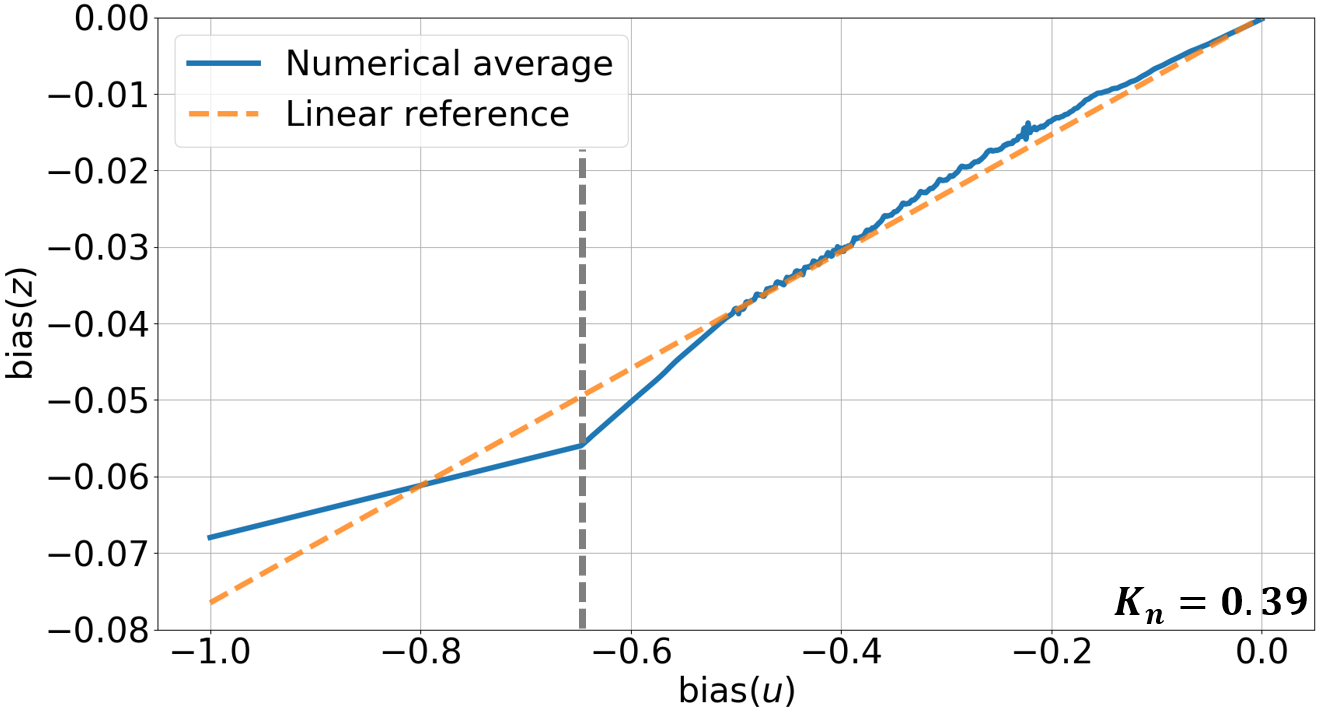}%
\label{fig:bias39}}\\
\subfloat[]{\includegraphics[width=0.79\columnwidth]{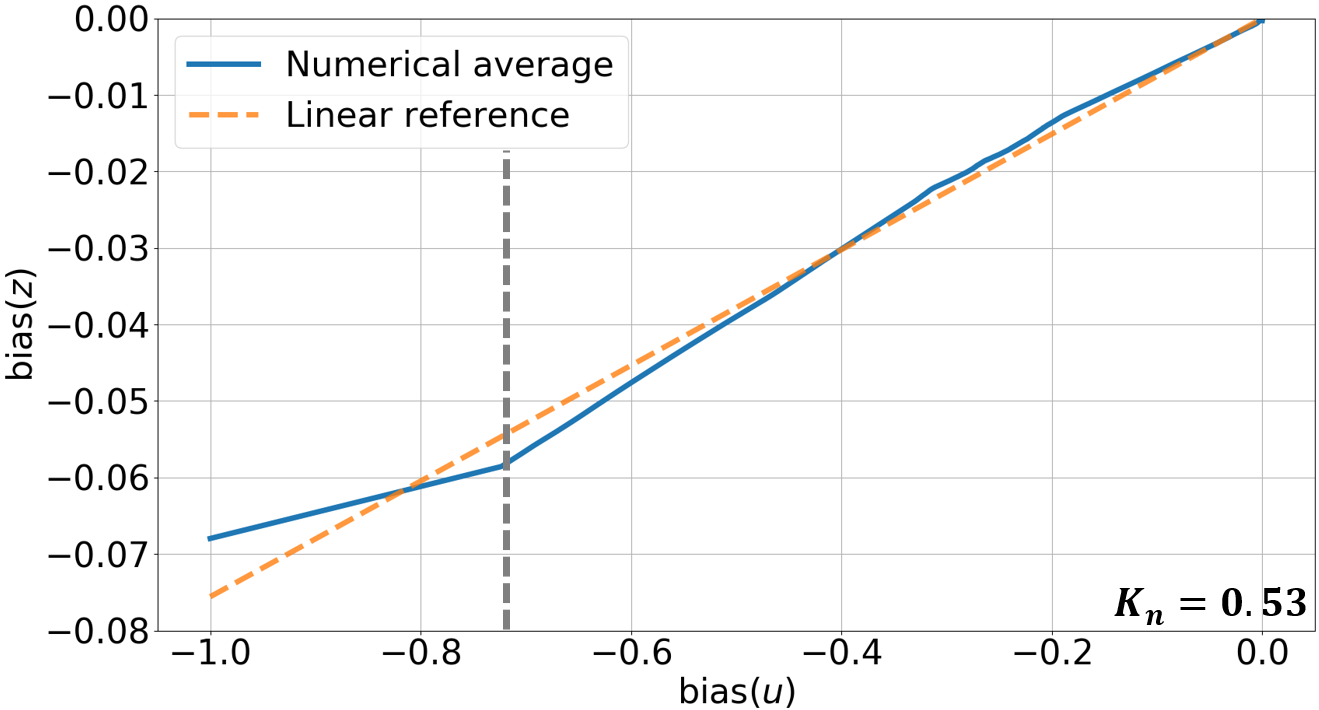}%
\label{fig:bias53}}\\
\subfloat[]{\includegraphics[width=0.79\columnwidth]{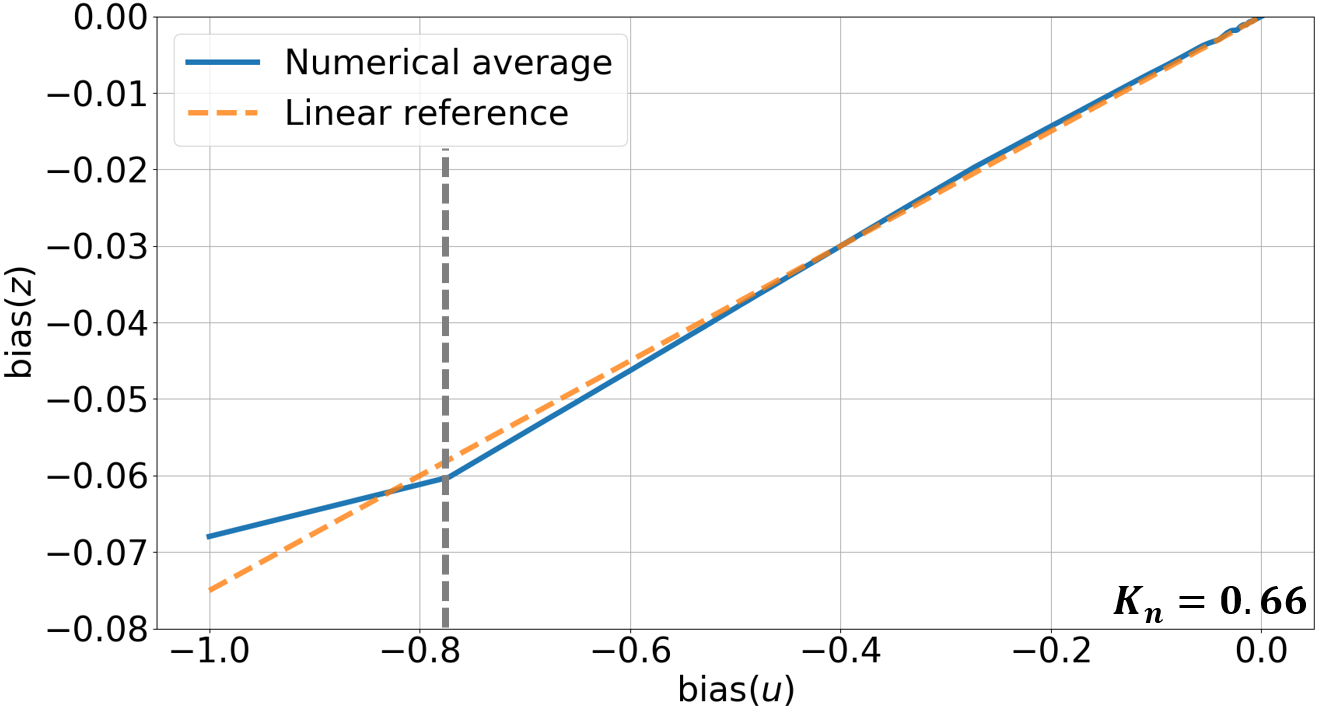}%
\label{fig:bias66}}\\
\subfloat[]{\includegraphics[width=0.79\columnwidth]{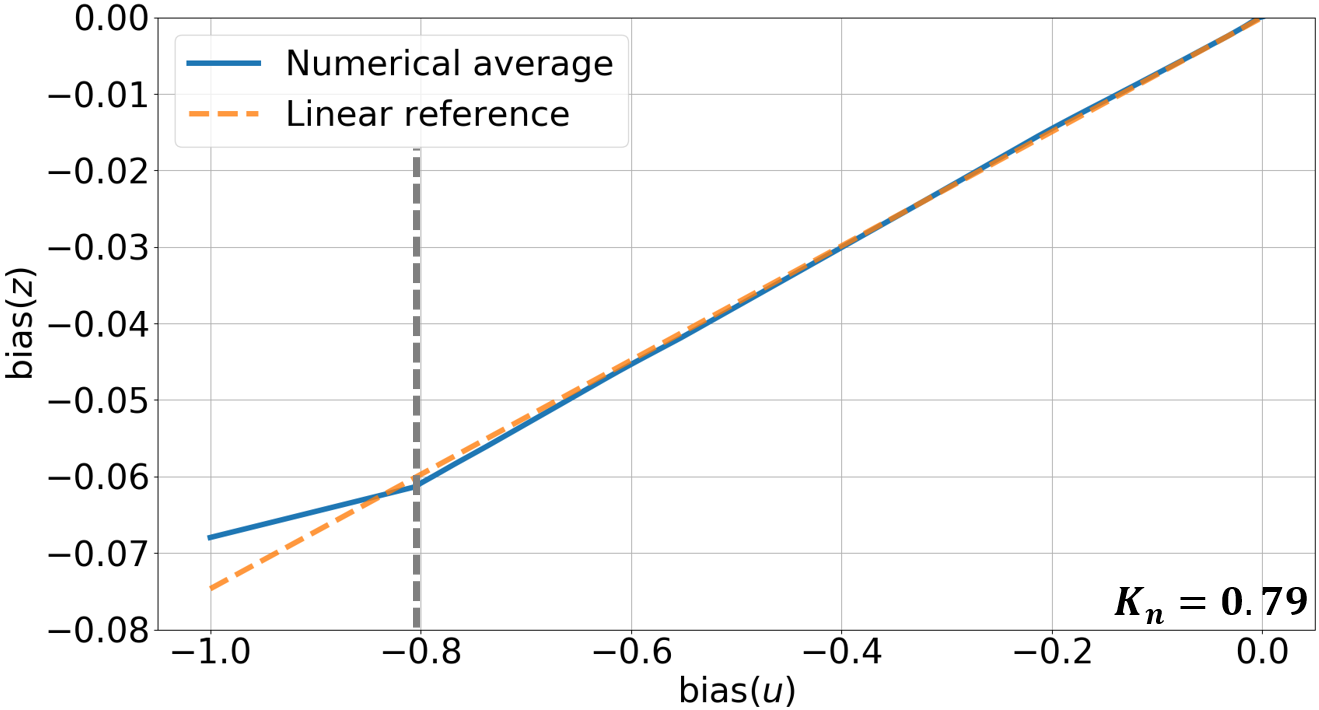}%
\label{fig:bias79}}
\caption{Relation between oscillation bias and extensor tonic input $u^e$ when setting different $a$ values to obtain (a) $K_n=0.19$ (b) $K_n = 0.39$ (a) $K_n=0.53$ (b) $K_n = 0.66$ (a) $K_n=0.79$. }
\label{fig:bias}
\end{figure}



\subsection{Steering control with imbalanced tonic inputs}
\label{sec:steer}

Most existing methods based on CPG realize steering by either directly adding a displacement \cite{ijspeert2008central} to the output of the CPG system, or using a secondary system such as an artificial neural network to compose the weighted outputs from multiple CPG systems \cite{mori2004reinforcement}. In this section, we present a different approach based on the maneuverability of the Matsuoka oscillator--tuning tonic inputs to realize the biased wave patterns of CPG outputs for steering the slithering locomotion of the soft snake robot\footnote{The fact that the biased wave output of the Matsuoka CPG system could cause the turning behavior of the snake robot comes from a previous work \cite{6420917}, which shows that the steering angle of a slithering snake robot on the planner ground can be linearly controlled by the bias of the oscillatory output of the Matsuoka oscillator as the command signal of joint actuators.}.

For the RL controller to steer our snake robot smoothly through the Matsuoka CPG system, we need to find a clever way to make the steering dynamics easy to learn for the RL algorithm. In other words, the relation between tonic inputs and the output bias of each primitive Matsuoka oscillator in the CPG network needs to be simple and clear. In the original design of the Matsuoka oscillator, the flexor and extensor tonic inputs are independent of each other. This setting not only increase the dimension of action space for the RL agent but also makes the relationship between tonic inputs and the output bias more complicated. To simplify this problem, we first introduce a new relation defined as \textit{complementation} to reform the relation between $u^e$ and $u^f$.

\begin{definition}
\label{def:complementary}
(Complementation) For two real signals $u(t)$ and $v(t)$, and a known bounded range $\calD:[a, b]$ where $\calD \subseteq \reals$, we say $u(t)$ and $v(t)$ are complementary to each other in range $\calD$ when $u(t), v(t) \in \calD$ for all $t\in \reals^+$ and $u(t)+v(t) \equiv b-a$.
\end{definition}

Another important definition for this section is a relation between two periodic signals named \textit{entrainment} based on the related theory in \cite{matsuoka2011analysis, Matsuoka2013FrequencyRO}.

\begin{definition}
\label{def:entrainment}
(Entrainment) Given a neural oscillator system with its natural frequency $\omega_n>0$. If the neural oscillator's output is synchronized to the coupled input with frequency $\omega$, then this system is entrained with the coupled input signal. The relation between the neural oscillator's output and the coupled input signal is called \textit{entrainment}. If the two signals are \textit{perfectly entrained}, they are supposed to have the same oscillation amplitude and bias in addition to the synchronized oscillation frequency.
\end{definition}

From our previous work \cite{xliu2020}, we have observed in experiment that the steering bias of a primitive Matsuoka oscillator is proportional to the amplitude of $u^e$ when $u^e$ and $u^f$ are complementary within the range $[0, 1]$. This key observation inspires us a dimension reduction technique to the input space of the CPG net: Instead of controlling $u^e_i, u^f_i$ for $i=1,\ldots, n$ for a $n$-link snake robot, we only need to control $u^e_i$ for $i=1,\ldots, n$ and let  $u^f_i=1-u^e_i$.
As the tonic inputs have to be positive in Matsuoka oscillators, we define a four dimensional action vector $\bm{\alpha}= [\alpha_1, \alpha_2, \alpha_3, \alpha_4]^T \in \reals^4$ and map $\bm{\alpha}$ to tonic input vector $\vec{u}$ as follows,
\begin{equation}
\label{eq:decoder}
u_i^e = \frac{1}{1+e^{-\alpha_i}},\text{ and } 
               u_i^f =1-u_i^e, \text{ for } i=1,\ldots, 4.
\end{equation}
 This mapping bounds the tonic input within $[0, 1]$. 
The reduced input dimension enables a more efficient policy search in RL. 

Based on this design, we show that there are certain combinations of tonic inputs in a Matsuoka oscillator that are capable of generating imbalanced output trajectories and therefore result in the turning behavior of the robot. We present three possible cases of the forced tonic inputs that could maneuver the turning behavior of the snake robot: 
\begin{enumerate}
    \item The two tonic inputs are different constants.
    \item The two tonic inputs are wave functions with imbalanced duty cycles. 
    \item The two tonic inputs are wave functions with imbalanced duty cycles, and both wave functions are added by different constant offsets. 
\end{enumerate}
It is noted that the third case is a linear combination of the first two. As a result, as long as the first two cases are proved to share the same property, the third one naturally holds. Next, we provide the frequency domain analysis of the Matsuoka oscillator to explain why the first two cases of tonic inputs enable imbalanced oscillation for the turning behavior.

\noindent \textbf{Steering with biased amplitude of constant tonic inputs: } To show that a pair of constant tonic inputs with different bias values can result in a biased oscillating output trajectory, we need to find out the relation between the bias of the output $z$ and the bias of tonic inputs, when the tonic inputs are constant and complementary in $[0,1]$. In this situation, a primitive Matsuoka oscillator needs to be a zero damping harmonic system to maintain limit cycle oscillation. When the system has zero damping, the ratio between the amplitudes of state $x^q$ and output $z^q$ for $q \in \{e, f\}$, referred to as $K_n$, is obtained from a second-order linear ordinary differential equation (\eqref{eq:odematsuoka} in Appendix~\ref{sec:kn}) derived from \eqref{eq:matsuoka}:
\begin{align}
	\label{def:Kn}
	K_n = \frac{\tau_r+\tau_a}{\tau_a a},
\end{align}

where $\tau_r$ and $\tau_a$ are the discharge rate and the adaptation rate in \eqref{eq:matsuoka}, and parameter $a$ is the mutual inhibition weight between flexor and extensor of a primitive Matsuoka oscillator. The derivation of \eqref{def:Kn} can be found in Appendix~\ref{sec:kn}.
 
When the Matsuoka oscillator's output only consists of free-response oscillation, we can establish  the following relation between the output $\text{bias}(z)$ and the tonic input $\text{bias}(u)$.

\begin{proposition}
  \label{prop:matsuokabias}
 If a primitive Matsuoka oscillator satisfies the following three conditions:
    \begin{inparaenum}[1)]
        \item the dynamical model of the primitive Matsuoka oscillator is harmonic, 
        \item the tonic inputs $u^e$ and $u^f$ are constants and complementary to each other,
        \item states $x^e$ and $x^f$ are perfectly entrained,
    \end{inparaenum} 
    then the oscillation bias of outputs $z$ and the bias of inputs $u$ satisfies the following linear relationship, 
    \begin{equation}
    \label{eq:amplitudebias}
        \mbox{bias}(z) = \frac{K_n}{(b-a)K_n + 1} \text{bias}(u),
      \end{equation}
      where $z=z^e-z^f$, $u = u^e-u^f$, and the coefficient $K_n$ satisfies $K_n = (\tau_r + \tau_a)/(\tau_a a)$.
\end{proposition}

\begin{IEEEproof}
See Appendix~\ref{sec:prop1}.
\end{IEEEproof}

Equation~\eqref{eq:amplitudebias} suggests that there is a linear relationship between $\mbox{bias}(u)$ and  $\mbox{bias}(z)$ in a primitive Matsuoka oscillator. We further validate this conclusion through the numerical simulation of a single primitive Matsuoka oscillator. We calculate the mean oscillation bias (numerical average\footnote{Based on Fourier series analysis, given a continuous real-valued $P$-periodic function $z(t)$, the  constant  term of its Fourier series has the form $\frac{1}{P}\int_P z(t) dt.$}) of the simulated state output $z$ and compare it with the estimated bias based on \eqref{eq:amplitudebias} (linear reference). Figure~\ref{fig:bias} shows the curve of $\mbox{bias}(z)$ varies with $\mbox{bias}(u)\in[-1,1]$ in a primitive Matsuoka oscillator. 

\begin{figure}[h!]
    \centering
    \includegraphics[width=0.6\columnwidth]{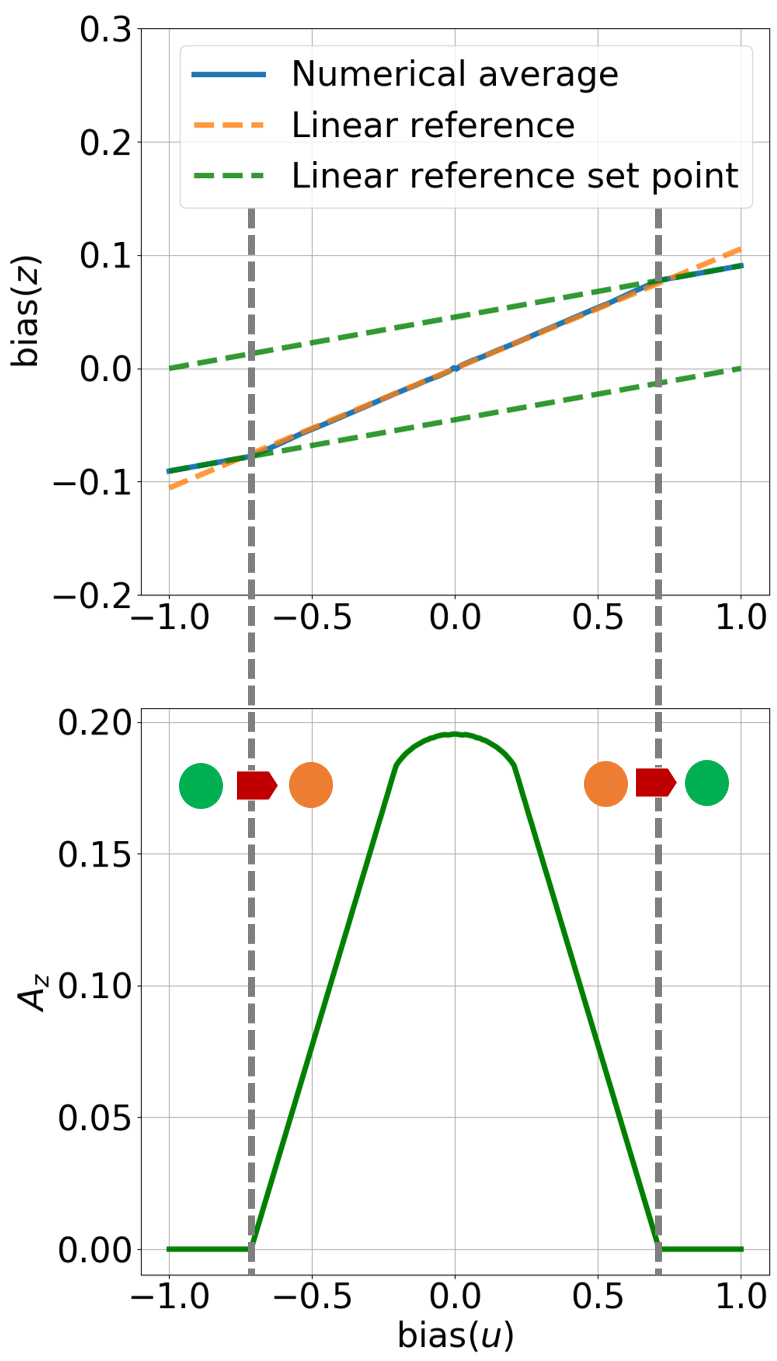}
    \caption{Relation between oscillation amplitude and duty cycle bias.}
    \label{fig:sp_lc_trans}
\end{figure}

Figure~\ref{fig:bias} and theoretical analysis (see Appendix~\ref{sec:prop1limit}) show that for $\mbox{bias}(u) \in (\frac{2a}{a+b+1}-1,1-\frac{2a}{a+b+1})$, the linear relationship mentioned in Proposition~\ref{prop:matsuokabias} is applicable to the data of $\mbox{bias}(z)$ and $\mbox{bias}(u)$ collected by simulating the original Matsuoka system in \eqref{eq:matsuoka}. It is also observed that the applicable range of $\mbox{bias}(u)$ for Proposition~\ref{prop:matsuokabias} to hold increases with $K_n$. As shown in Fig.~\ref{fig:sp_lc_trans}, when $\mbox{bias}(u) \in [-1,\frac{2a}{a+b+1}-1]\cup[1-\frac{2a}{a+b+1},1]$, the original Matsuoka system stops oscillating, which means the system stays at a set point equilibrium. In this case, $\mbox{bias}(z)$ and $\mbox{bias}(u)$ follow another linear relationship,
\begin{align}
    \mbox{bias}(z) = \begin{cases}
    \frac{\mbox{bias}(u) - (1+2c)}{2(1+b)}, \quad \text{if } \mbox{bias}(u) \in [-1,\frac{2a}{a+b+1}-1]\\
    \frac{\mbox{bias}(u) + (1+2c)}{2(1+b)}, \quad \text{if } \mbox{bias}(u) \in [1-\frac{2a}{a+b+1},1].
    \end{cases}
\end{align}
The derivation of the above relationship is provided in Appendix~\ref{sec:prop1limit}.

In the next paragraph, we show that there is also a linear relationship between $\mbox{bias}(z)$ and $\mbox{bias}(u)$ of the Matsuoka oscillator when $u^e$ and $u^f$ are periodical signals with biased duty cycles.

\noindent \textbf{Steering with the biased duty cycle of periodic tonic inputs: } We show a different approach to control the steering of the snake robot given that both $u_i^e$ and $u_i^f$ are square wave functions and are complementary to each other. 

\begin{proposition}
\label{prop:matsuokaduty}
    If a primitive Matsuoka oscillator satisfies the following three conditions:
    \begin{inparaenum}[1)]
        \item the dynamical model of the primitive Matsuoka oscillator is harmonic, 
        \item the tonic inputs $u^e$ and $u^f$ are square wave signals and are complementary to each other, 
        \item $u^e$ is entrained with $z^e$, and $u^f$ is entrained with $z^f$,
    \end{inparaenum}
    then the oscillation bias of $z$ and the bias of $u$ satisfies the following linear relationship, 
    \begin{align}
    \label{eq:dutybias}
    \mbox{bias}(z) = \frac{1+2 m}{b-a+2}\mbox{bias}(u),
    \end{align}
    where  $z=z^e-z^f$, $u = u^e-u^f$, and 
    \[
        m = \frac{1}{\pi}\frac{1}{2K_n-1+\frac{2}{\pi}(a+b)\sin^{-1}(K_n)}
    \]
     is a constant coefficient ($r$ indicates amplitude of state $x$). 
\end{proposition}
\begin{IEEEproof}
  See Appendix~\ref{sec:prop2}.
  \end{IEEEproof}

\begin{figure}[h!]
\centering
\subfloat[]{\includegraphics[width=0.82\columnwidth]{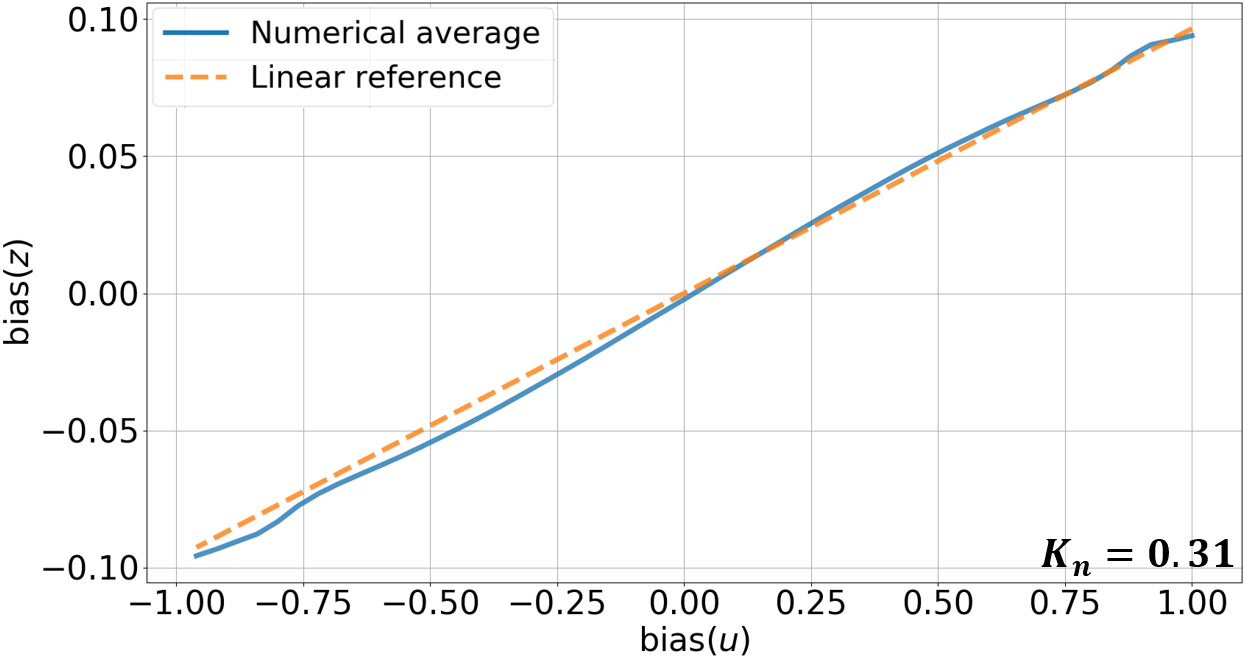}%
\label{fig:duty31}}\\ 
\subfloat[]{\includegraphics[width=0.82\columnwidth]{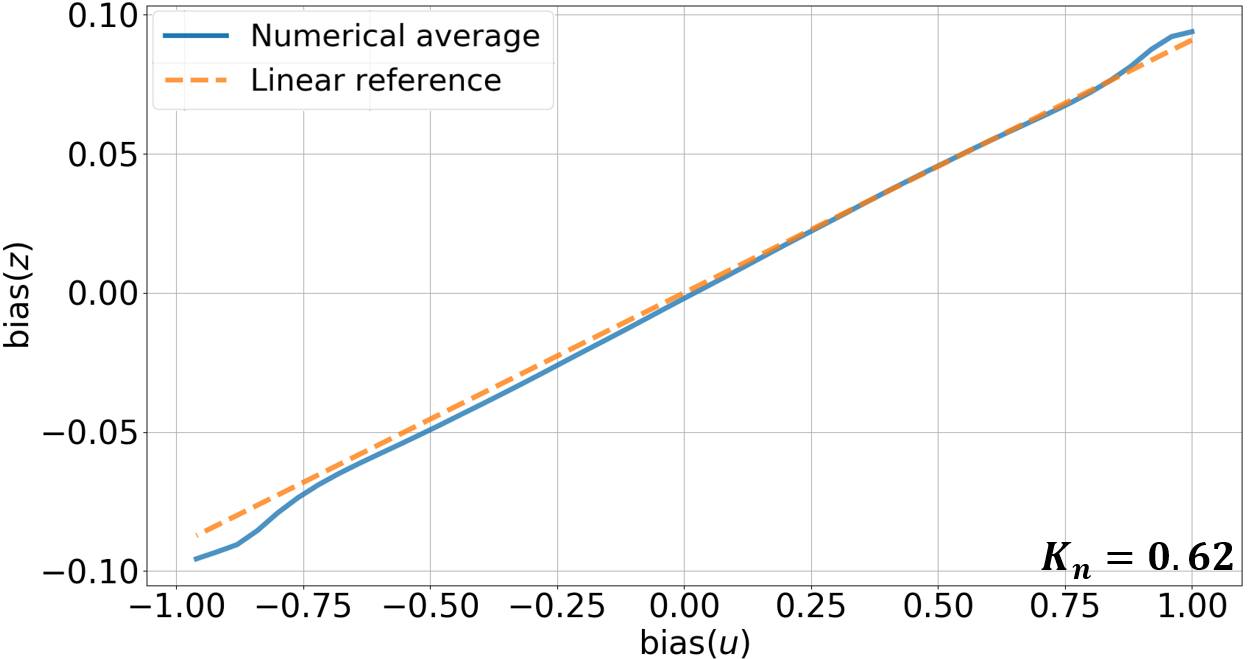}%
\label{fig:duty62}}\\
\subfloat[]{\includegraphics[width=0.82\columnwidth]{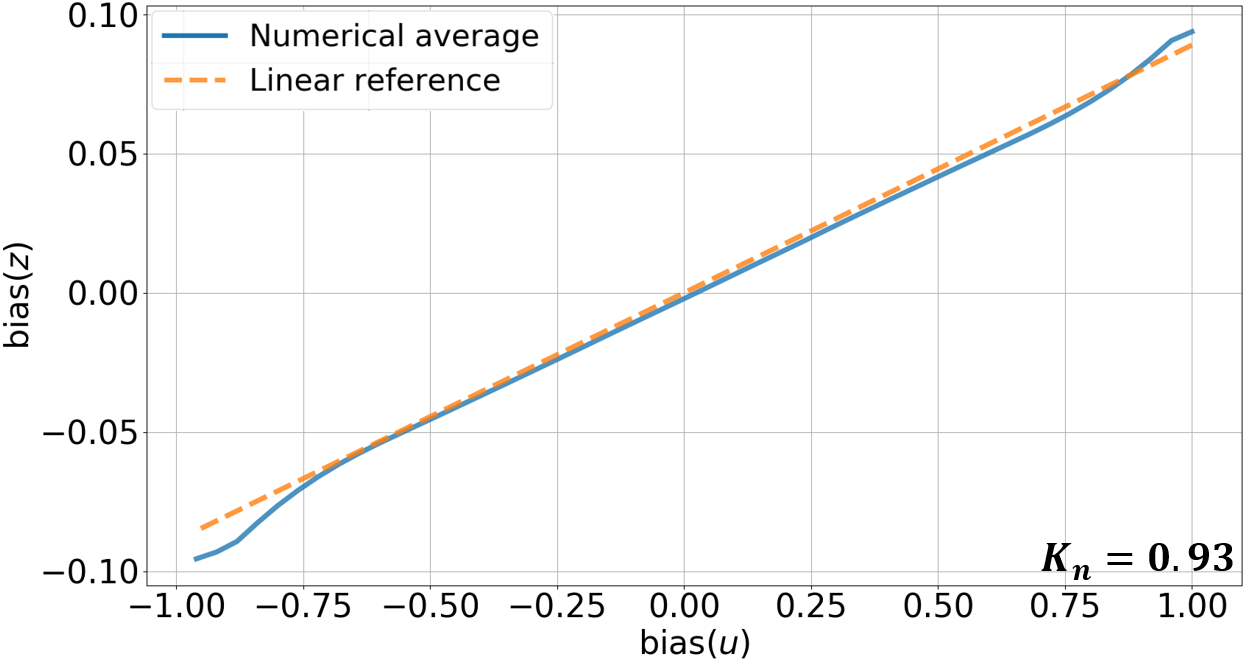}%
\label{fig:duty93}}
\caption{Relation between $\mbox{bias}(z)$ and $\mbox{bias}(u)$ for the tonic inputs satisfying Proposition~\ref{prop:matsuokaduty}.}
\label{fig:dutybias}
\end{figure}


The simulated results also supports Proposition~\ref{prop:matsuokaduty} when the Matsuoka system is taking periodic tonic inputs with biased duty cycles. Figure~\ref{fig:dutybias} shows that with various $K_n$ values, the linear relationship in \eqref{eq:dutybias} fits well with the curve between $\mbox{bias}(u)$ and $\mbox{bias}(z)$ collected by simulating the original Matsuoka system in \eqref{eq:matsuoka}. 

Combining the conclusions in Proposition~\ref{prop:matsuokabias} and Proposition~\ref{prop:matsuokaduty}, we make the following remark,

\begin{remark}
\label{remark:bias}
If a primitive Matsuoka oscillator has periodical tonic input signals $u^e$ and $u^f$ that are complementary to each other, with imbalanced duty cycles and both wave functions are added by different constant offsets, then $\mbox{bias}(z)$ is linearly related to the $\mbox{bias}(u)$, where  $z=z^e-z^f$, $u = u^e-u^f$.
\end{remark}

 Proposition~\ref{prop:matsuokabias} and \ref{prop:matsuokaduty} show that the oscillation bias of the Matsuoka CPG system is easy to maneuver through the biased tonic input signals. Since the oscillation bias is the key to steering in the snake's slithering locomotion,  these two propositions provide us insight to the design of RL module  so as to improve the efficiency in learning the steering behavior of the snake robot.
 
\subsection{Velocity control with frequency modulation}
\label{sec:vel}

\begin{figure}[h!]
    \centering
    \includegraphics[width=0.85\columnwidth]{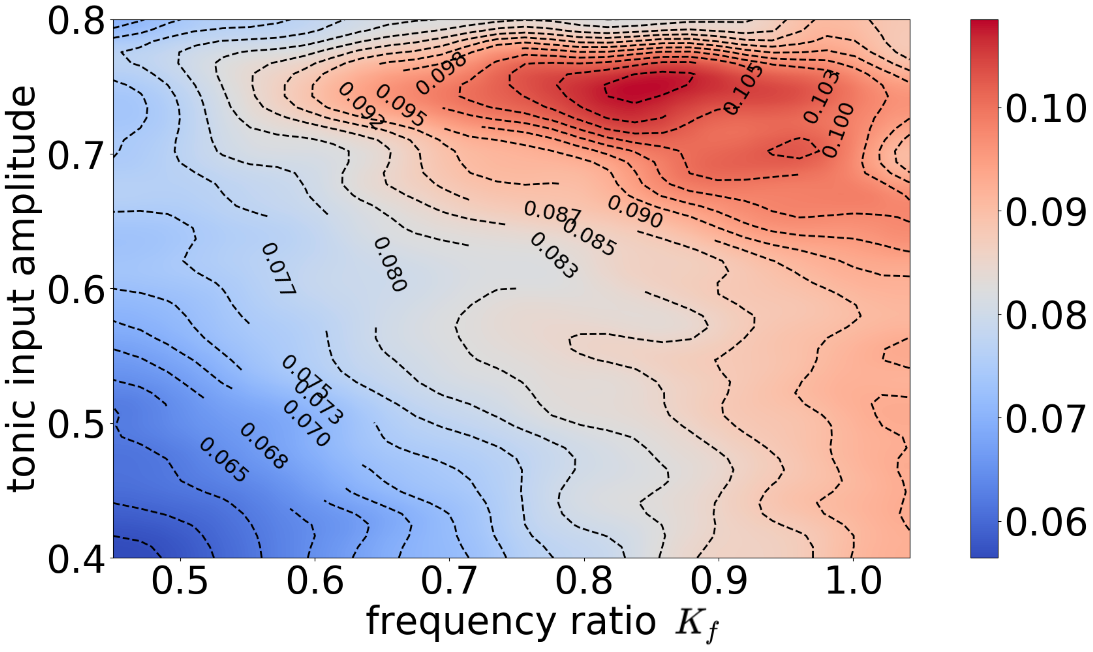}
    \caption{Relating oscillating frequency and amplitude to the average linear velocity of serpentine locomotion.}
    \label{fig:ampfreqvel}
\end{figure}
 
Generally, the linear velocity of serpentine locomotion is affected by the snake's oscillation amplitude and frequency. In this subsection, we show that the amplitude and frequency can be controlled by two coefficients of the Matsuoka CPG system to change the locomotion velocity of the soft snake robot.

First, the following relation between the frequency ratio $K_f$ and the natural frequency $\hat \omega_i$ of the $i$-th oscillator is established in \cite[(5),(6)]{matsuoka2011analysis},
\begin{align}
\label{eq:kf_property}
      \hat{\omega_i} \propto \frac{1}{\sqrt{K_f}},  i \in \{1, 2, 3, 4\}.
\end{align}

Second, the oscillating amplitude $\hat{A_i}$ of the $i$-th oscillator is linearly proportional to the amplitude of free-response oscillation tonic input $c$ when $c>0$ and $u_i^e, u_i^f$ are constants \cite{matsuoka2011analysis}, that is, 
\begin{equation}
    \label{eq:amplitude}
    \hat{A_i} \propto c,  i \in \{1, 2, 3, 4\}.
\end{equation}

Equations \eqref{eq:kf_property} and \eqref{eq:amplitude} show that the frequency and amplitude of the Matsuoka CPG system are \emph{independently} influenced by the frequency ratio $K_f$ and the free-response oscillation tonic inputs $c$. Therefore, these two coefficients can be considered major factors for the Matsuoka CPG system to control the velocity of the soft snake robot's locomotion. In
Fig.~\ref{fig:ampfreqvel}, we collect $2500$ uniform samples within the region $c \in [0.4, 0.8]$, and $K_f \in [0.45, 1.05]$ and record the velocities generated in the simulator. We observe that with a fixed $c$, the average velocity increase monotonically with the frequency ratio $K_f$. We also observe that with the same $K_f$, the change of $c$ does not affect the locomotion velocity significantly. While with different values of $c$, the efficiency of $K_f$ in affecting the locomotion velocity is different. This means that we can mainly use $K_f$ to adjust the locomotion velocity, but the value of $c$ needs to be carefully selected. Given this analysis, we use $K_f$ to control the velocity of the robot. It is noted that the frequency ratio $K_f$ only influences the strength but not the direction of the vector field of the Matsuoka CPG system. Thus, modulating $K_f$ would not affect the stability of the whole CPG system. 

\subsection{Modulating forced-response oscillation amplitude with free-response oscillation tonic input constraint}
\label{sec:FOC}

\eqref{eq:amplitude} shows that the free-response oscillation tonic input $c$ could affect the output amplitude of the Matsuoka oscillator when $u^e$ and $u^f$ are constants. We further discover that a positive value of the free response tonic input $c$ could set a threshold for the amplitude of the force-response tonic inputs $u^e$ and $u^f$, such that they need to pass this amplitude threshold in order to control the oscillation of the CPG system. In the experiment section, we show that this property of $c$ can significantly improve the sim-to-real performance of our control framework. 

In our previous work\cite{xliu2020}, when $c=0$, there is no free-response oscillation in the system. When a Matsuoka oscillator has no free-response oscillation pattern, its output oscillation amplitude and bias are only controlled by the forced input signal given by the control tonic inputs $u^e$ and $u^f$. When the inertia in the simulated learning environment is high and the contact friction force is low, the RL agent learns to generate the forced-response oscillation tonic inputs with very small amplitude to keep a more stable heading direction during the locomotion. However, if we need the RL control policy to be able to initiate the CPG oscillation with an increased amplitude on the real robot (e.g. for traversing a terrain with higher friction resistance), the learned policy would not meet the requirement.

When $c\neq 0$, we conclude that in the Matsuoka oscillator, the amplitude $A_u$ of the force-response tonic inputs $u^e$ and $u^f$ must satisfy the inequality $A_u > A_0$ to completely entrain with the output $z$ ($A_0$ is the entrainment threshold for $u^e$ and $u^f$ to synchronize the output $z$ of the Matsuoka oscillator \cite{Matsuoka2013FrequencyRO}). The equation of $A_0$ is given as follows

\begin{equation}
    \label{eq:A0}
    A_0(c,\omega) = \frac{c}{\frac{\sqrt{\tau_a^2\omega^2 + 1}}{\tau_r\tau_a|\omega_n^2 - \omega^2|} \frac{c+1}{A_n}},
\end{equation}
where $A_n >0$ is the free-response oscillation amplitude and 
\[
\omega_n = \frac{1}{\tau_a K_f} \sqrt{\frac{(\tau_r+\tau_a)b}{\tau_r a}-1}
\]
is the free-response oscillation frequency \cite{matsuoka2011analysis}. The detailed derivation of $A_0$ is provided in Appendix~\ref{sec:amp}. According to \cite[(30)]{matsuoka2011analysis}, we have
\begin{align}
\label{eq:a0approx}
    A_0(c,\omega) \approx \frac{c}{\frac{\sqrt{\tau_a^2\omega^2 + 1}}{\tau_r\tau_a|\omega_n^2 - \omega^2|} (2K_n-1+\frac{2}{\pi}(a+b)\sin^{-1}(K_n))}.
\end{align}

In \eqref{eq:a0approx}, if $c=0$, $A_0 \equiv 0$. In this case, there is no limiting threshold for the control policy to entrain the CPG output $z$ with $u^e, u^f$. When $\omega$ is fixed and $c>0$, then the threshold $A_0 >0$ and $A_0$ increases with $c$. Notice that $A_u>A_0$ must be satisfied for the free-response oscillation of the Matsuoka system be attenuated by the system damping. This also means the force-response tonic inputs $u^e, u^f$ entrain the CPG output $z$. In this scenario, the control policy needs to increase $A_u$ to control the CPG system effectively. It is also noted that $A_0 \rightarrow 0$ as $\omega \rightarrow \omega_n$, therefore there are two ways for the RL agent to realize the entrainment status: one is keeping the oscillation frequency $\omega$ close to the free-response oscillation frequency $\omega_n$, and the other is increasing the value of $A_u$ to make $A_u>A_0$. Therefore, the combination of the two directions can encourage the intelligent controller to produce force-response tonic inputs that can not only approach desired oscillating amplitude, but also pursue frequency resonance with the original CPG system. Based on this special property of the Matsuoka oscillator, we propose a new method -- FOC-PPOC-CPG to enforce better entrainment between RL control signals and the CPG states.

 According to the relation between $A_0$ and $c$, we can use $c$ to keep the oscillation amplitude of the Matsuoka oscillator at different levels.  One previous work \cite{wang2017central} has shown that the oscillation amplitude of the Matsuoka oscillator can be used to improve the slithering locomotion velocity of a rigid snake robot in different environments with different friction coefficients. Hence, we can use $c$ to adapt the body undulation amplitude of the soft snake robot to different environments with various contact properties. With this approach, we can improve the sim-to-real performance of an RL snake controller by tuning its signal amplitude, instead of relying on the environment-based methods such as domain randomization \cite{tobin2017domain} or other data augmentation techniques, which are computationally expensive.

In the later part of this paper, our experiment results (see Section~\ref{sec:focsin2real}) verify the merit of $c$ in improving the sim-to-real performance of our snake locomotion controller. 

\subsection{The Neural Network Controller}
We have now determined the encoded input vector of the CPG net to be vector $\bm{\alpha}$ (tonic inputs) and frequency ratio $K_f$. This input vector of the CPG is the output vector of the NN controller. The input to the NN controller is the state feedback of the robot, given by $s=[||\bm{\rho}_g||, v_g, \theta_g, \Dot{\theta_g}, \kappa_1, \kappa_2, \kappa_3, \kappa_4]^T \in \reals^8$ (see Fig.~\ref{fig:coordinate}). Next, we present the design of the NN controller.

The key insight for the design of the NN controller is that the robot needs not to change velocity very often for smooth locomotion. This means the updates for tonic inputs and the frequency ratio can be set to be at two different time scales. With this insight, 
we adopt a hierarchical reinforcement learning method called the option framework \cite{sutton1999between,precup2000temporal} to learn the optimal controller. The controller uses the tonic inputs as low-level primitive actions and frequency ratio as high-level options of the CPG net. The low-level primitive actions are computed at every time step. The high-level option changes infrequently. Specifically, each option is defined by $\langle {\cal{I}}, \pi_y: S\rightarrow \{y\} \times \reals^4), \beta_{y} \rangle $ where ${\cal I}= S$ is a set of initial states, and $\pi_y$ is the intra-option policy. By letting ${\cal I}=S$, we allow the frequency ratio to be changed at any state in the system. Variable $y\in [0,1]$ is a value of frequency ratio, and $\beta_y: S\rightarrow[0,1]$ is the termination function such that $\beta_y(s)$ is the probability of changing from the current frequency ratio to another frequency ratio.

The options share the same NN for their intro-option policies and the same NN for termination functions. However, these NNs for intro-option policies take different frequency ratios. The set of parameters to be learned by policy search include parameters for intra-option policy function approximation, parameters for termination function approximation, and parameters for high-level policy function approximation (for determining the next option/frequency ratio). Proximal Policy Optimization Option-Critics (PPOC) in the OpenAI Baselines \cite{baselines} is employed as the policy search in the RL module.

Let us now review the control architecture in Figure~\ref{fig:rlmatsuoka}. We have a Multi-layer perceptron (MLP) neural network with two hidden layers to approximate the optimal control policy that controls the inputs of the CPG net in \eqref{eq:matsuoka}. The output layer of MLP is composed of action $\bm{\alpha}$ (green nodes), option in terms of frequency ratio (pink node), and the terminating probability (blue node) for that option. The input of NN consists of a state vector (yellow nodes) and its output from the last time step. The purpose of this design is to let the actor network learn the unknown dynamics of the system by tracking the past actions in one or multiple steps \cite{mori2004reinforcement, peng2018sim, hwangbo2019learning}.
Given the Bounded Input Bounded Output (BIBO) stability of the Matsuoka CPG net \cite{matsuoka1985sustained} and that of the soft snake robots, we ensure that the closed-loop robot system with the FOC-PPOC-CPG controller is BIBO stable. Combining with \eqref{eq:decoder} which enforces a limited range for all tonic inputs, this control scheme is guaranteed to generate bounded inputs, which lead to bounded outputs in the system.

\section{Curriculum and Reward Design for Efficient Learning-based Control}
\label{sec:sol}

In this section, we introduce  the design of the curriculum and reward function for efficiently learning a goal-tracking controller given the proposed FOC-PPOC-CPG scheme.

\subsection{Task curriculum}
\label{sec:curriculum}

\begin{figure}[h!]
    \centering
    \includegraphics[width=0.85\columnwidth]{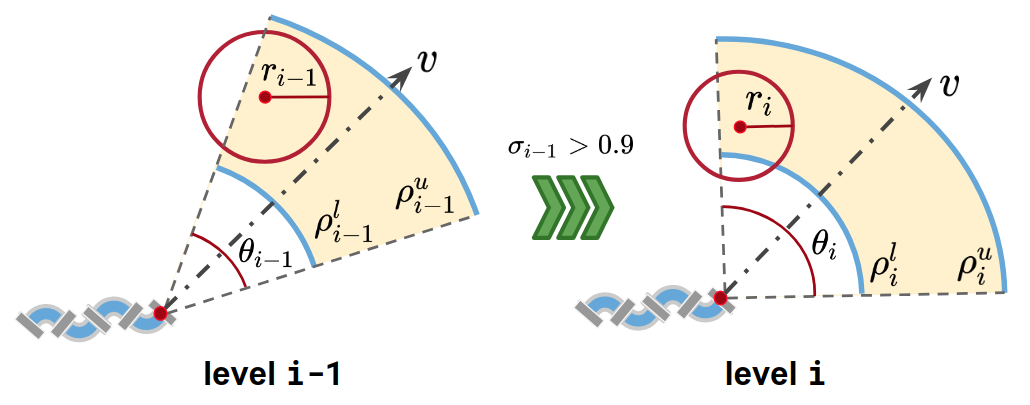}
    \caption{Task difficulty upgrade from level $i-1$ to level $i$. As the curriculum level increases, goals are sampled at a narrower distance and wider angle, and the acceptance area gets smaller.}
    \label{fig:curriculumlvlup}
\end{figure} 

 Curriculum teaching \cite{karpathy2012curriculum} is used to accelerate motor skills learning given complex goal-tracking tasks. We design the curriculum such that the agent starts with easy-to-reach goals at level 0. As the level increases, the agent learns to perform more challenging goal-tracking tasks.

The curriculum levels are designed as follows: At the task-level, $i$, the center of the goal is sampled from the 2D space based on the current location and head direction of the robot. For each sampled goal, we say the robot reaches the goal if it is $r_{i}$ distance away from the goal. The sampling distribution is uniform in the fan area determined by the range of angle $\theta_i$ and distance bound $[\rho_i^l, \rho_i^u]$ in the polar coordinate given by the predefined curriculum.
 
As shown in Fig.~\ref{fig:curriculumlvlup}, when the task-level increases, we have $r_{i}<r_{i-1}$, $\theta_i> \theta_{i-1}$,  $\rho_i^u >\rho_{i-1}^{u}$, and $\rho_i^u-\rho_i^l < \rho_{i-1}^u-\rho_{i-1}^l$. This means that the robot has to be closer to the goal in order to succeed and receive a terminal reward, the goal is sampled in a range further from the initial position of the robot.  We select discrete sets of $\{r_i\}, \{\theta_i\}$, $[\rho_i^{l},\rho_i^{u}]$ and determine a curriculum table. A detailed example of the learning curriculum is given in Table~\ref{tab:curriculum}. We train the robot in simulation starting from level $0$. The task-level is increased to level $i+1$ from level $i$ if the controller reaches the desired success rate $\sigma_i$, for example, $\sigma_i=0.9$ indicates at least $90$ successful completions of goal-reaching tasks out of $n=100$ trials at level $i$.  

\subsection{Reward design}

The design of the reward function is to guide the robot to the set point goals. We consider building the artificial potential field \cite{choset2005principles} such that the robot is attracted by the goal $g$. We use a simple conical potential field for each goal. For any position represented by coordinate $\mathbf{x}$ in Cartesian space, let vector $\mathbf{e}_g = \mathbf{x}_g - \mathbf{x}$, the norm $||\mathbf{e}_g||$ indicates the distance between the position of the robot's head and the goal. The constant attracting force at $\mathbf{x}$ becomes
\[
    \mathbf{f}_g(\mathbf{x}) = \frac{\mathbf{e}_g}{||\mathbf{e}_g||}. 
\]
Given the single goal-tracking scenario without obstacles, we have the potential field reward for goal-tracking as
\[
    U(\mathbf{x}) = \frac{\mathbf{v}_s \cdot \mathbf{f}_g(\mathbf{x})}{||\mathbf{e}_g||},
\]
where $\mathbf{v}_s$ is the velocity vector of the soft snake robot.

Combining with the definition of goal-reaching tasks and their corresponding level setups, the reward at every time step is defined as 

\begin{equation}
    R(v_g,\theta_g) = c_v v_g + c_g U +  c_g \cos \theta_g \sum_{k=0}^i{\frac{1}{r_k} I(||\bm{\rho_g}|| < r_k)},
\end{equation}
where $c_v, c_g \in \reals^+$ are constant weights, $v_g$ is the length of the projection of the snake's head COM velocity $\bm{v}$ on the head-to-goal-direction, $\bm{\rho}_g$ is the linear displacement vector between the head COM of the robot and the goal position, $\theta_g$ is the angle between vector $\bm{v}$ to vector $\bm{\rho}_g$ in Fig.~\ref{fig:coordinate}, $r_k$ defines the goal range in task-level $k$, for $k=0,\ldots, i$, and $I(\bm{\rho_g} < r_k)$ is an indicator function that outputs one if the robot head is within the goal range for task-level $k$. 

This reward trades off two objectives. The first term, weighted by $c_v$, encourages high locomotion velocity toward the goal. The second term, weighted by $c_g$, rewards the learner based on the position of the robot to the goal, and the level of the curriculum the learner has achieved for the goal-reaching task. For every task, if the robot hasn't entered the goal range, it receives a potential field reward only. When the robot enters the goal range in task-level $i$, it receives a summation of rewards $1/r_k$ for all $k\le i$ (the closer to the goal the higher this summation), shaped by the approaching angle $\theta_g$ (the closer the angle to zero, the higher the reward).

If the agent reaches the goal defined by the current task-level, a new goal is randomly sampled in the current or next level (if the current level is completed). 
There are two failing situations, where the desired goal is re-sampled and updated. The first situation is starving, which happens when the robot stops moving for a certain amount of time, referred to as the starvation time. The second case is missing the goal, which happens when the robot keeps heading in the wrong direction as opposed to moving towards the goal ($v_g(t)$ being negative) for a certain amount of time.

 \section{Experimental Evaluation}
\label{sec:result}

\begin{figure}[h!]
    \centering
    \includegraphics[width=0.85\columnwidth]{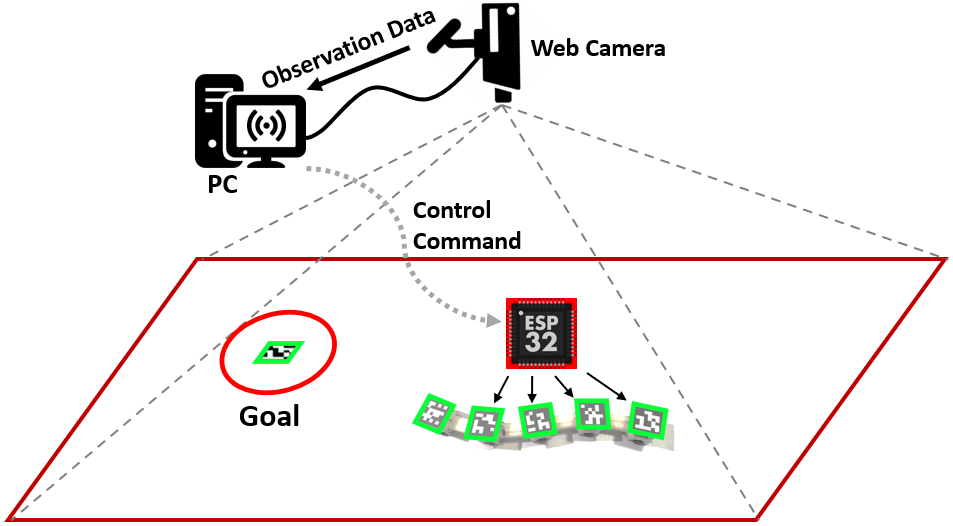}
    \caption{The currently used motion capture system for goal-tracking tasks.}
    \label{fig:mocap}

\end{figure}

In this section, we evaluate the proposed method in both simulation and real environments. We first introduce the experimental setup to explain how the data of the robot is collected during locomotion, as well as the training configuration for the RL algorithm. Then we compare the properties of control signals between our method and the vanilla PPO as locomotion controllers for goal-reaching tasks in simulation. In the comparison analysis, we highlight the performance drop of each method from simulation to real to extrapolate the advantage of our method. Last, we further test and analyze the sim-to-real robustness in difficult goal-reaching tasks that are never seen at the training stage, and the real robot performance against disturbance. 

\subsection{Experimental Setup}

\textbf{Environment Sensing and Data Collection:} The  states of the real snake robot are captured by a single web camera hanging on the ceiling of the experiment room. The robot body detection is realized by using Aruco -- a library in OpenCV for QR codes detection and localization \cite{GARRIDOJURADO20142280}. These QR codes are printed and attached to the rigid bodies of the snake robot and the goal position. Figure~\ref{fig:mocap} shows the experiment setup for the real snake robot goal-reaching tasks. Once the QR codes on the robot bodies and the goal(s) are detected, their pixel-wise coordinate vectors are calculated with distortion corrected. Given the camera calibration information, we can translate the pixel data of all QR codes into the real world 2D coordinates, and then transform it into positional information and the body posture of the robot. The control policy function running on the desktop computer receives the observation states, generates the control commands and passes them through WiFi communication. The ESP32 chips on the snake bodies translate the commands into Pulse Width Modulation (PWM) signals to activate or deactivate the valves\cite{luo2017toward, renato2019} on the snake robot. 

\textbf{Reinforcement Learning Configuration:} We use a four-layered NN with $128\times128$ hidden layer neurons as a general configuration for the actor and critic networks of all RL methods mentioned in this section. The back-propagation of the critic net was done with Adam Optimizer and a step size of $5\times10^{-4}$. For data collection of each trial trajectory, the starvation time for the failing condition is $60 \text{ ms}$. The missing goal criterion is triggered whenever $v_g(t)$ (the velocity on the goal-direction, see Fig.~\ref{fig:coordinate}) stays negative for over $60$ time steps. In order to compensate for the mismatch between the simulation and the real environment, most notably the friction coefficients, we employ a domain randomization technique \cite{tobin2017domain}, in which a subset of physical parameters are sampled from several uniform distributions. The range of distributions of domain randomization (DR) parameters used for training are in Table \ref{tab:dr} (see Appendix~\ref{sec:data}).

For PPOC-CPG and FOC-PPOC-CPG, we first train the policy net with fixed options (at this moment, the termination probability is always $0$, and a fixed frequency ratio $K_f= 1.0$ is used). When both the task-level and the reward cannot increase anymore, we allow the learning algorithm to change the option, \ie, pick a different frequency ratio $K_f$ along with termination function $\beta$, and keep training the policy until the highest level in the curriculum is passed. 

In the PPOC-CPG method, the value of the free-response tonic input $c$ is equivalently considered zero since it is not formally introduced in the previous control design \cite{xliu2020}. According to the definition of $A_0$ (\eqref{eq:A0}), the amplitudes of both $u^e$ and $u^f$ need to be greater than $A_0$ in order to dominate in controlling the outputs of the Matsuoka CPG system. The value of $A_0$ should not be greater than the upper bound of $u^e$ and $u^f$, which is $1$ defined by \eqref{eq:decoder}. Among a group of candidates ranging from $0.25$ to $2$, we choose $c=0.75$ as our free-response oscillation constraint for the FOC-PPOC-CPG controller. This value is valid for our system because when we set $c=0.75$ and all other coefficients of the CPG network (Table~\ref{tab:config}) to \eqref{eq:A0}, the result shows $A_0 \in [0.24, 0.34] \subset [0, 1]$, with $\omega \in [3.77, 5.02]$. It is noted that the range of $\omega$ here is calculated from multiple sampled sequences of $u^e$ and $u^f$ recorded in the real snake goal-reaching tasks. Since we are testing the sim-to-real performance, all methods involved in this comparison are trained in the simulator for sufficiently long iterations (12500 episodes) to ensure convergence. Each method is trained $10$ times with different random seeds and the controller with the best performance is selected to be tested on the real robot. All curriculum parameters (Table~\ref{tab:curriculum}) and domain randomization parameters (Table~\ref{tab:dr}) are fixed for all three methods involved.
 
 The whole training process of each method runs on $4$ simulated soft snake robots in parallel on a workstation equipped with an Intel Core i7-9700K, 32GB of RAM, and one NVIDIA RTX2080 Super GPU.

\subsection{Verification of steering property of PPOC-CPG}

\begin{figure}[h!]
\centering
\subfloat[]{\includegraphics[width=0.47\columnwidth]{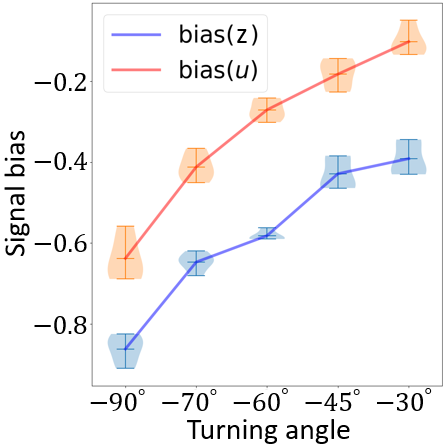}%
\label{fig:bias_compare}}  
\hfil
\subfloat[]{\includegraphics[width=0.45\columnwidth]{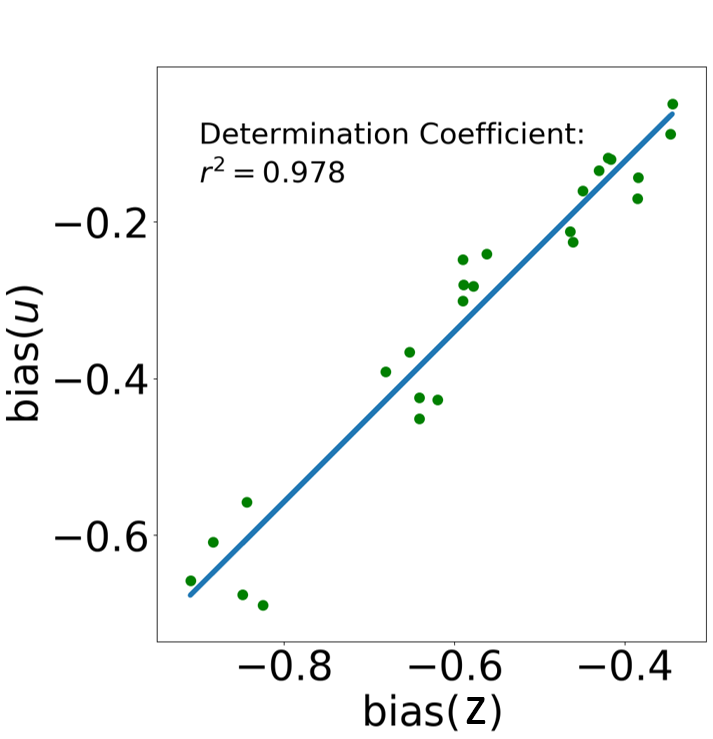}%
\label{fig:bias_regression}}
\caption{(a) Bias input and output of the RL-driven CPG node for different turning angles (mean values connected). (b) Linear relation between input and output bias of the RL-driven CPG node during locomotion.}
\label{fig:bias_test}
\end{figure}


We use a simulated experiment to show that our FOC-PPOC-CPG control policy has learned the turning behavior with the biased tonic input signal, and the Matsuoka CPG system can linearly map the biased tonic input to the biased actuation signal as Proposition~\ref{prop:matsuokabias} and Proposition~\ref{prop:matsuokaduty} predicted. In the experiment, we test the converged FOC-PPOC-CPG policy on multiple set-point goals placed in certain directions ($-90^\circ, -70^\circ, -60^\circ, -45^\circ, -30^\circ$) with a fixed distance ($1$ meter), which approximately represent the desired turning angles of the locomotion tasks. For each goal position, we carry out $5$ trials and record the tonic inputs data and CPG output data of the head CPG node of the soft snake robot. The reason for choosing the head node is because this node's behavior best reflects the desired steering direction of the RL agent. Figure~\ref{fig:bias_compare} shows a violin plot of the tonic input bias and the CPG output bias for different turning angles (the bias signals are calculated by \eqref{eq: biasu}). It is observed from Fig.~\ref{fig:bias_compare} that both bias signals are monotonically related to the desired turning angle (initial goal-direction). Figure~\ref{fig:bias_regression} shows the linear regression result based on all data points. We can observe a clear linear relationship between $\text{bias}(z)$ and $\text{bias}(u)$ of the head CPG node (with the coefficient of determination equal to $0.978$, a value closer to $1$ indicate higher linearity). This result provides stronger support for Proposition~\ref{prop:matsuokabias} and Proposition~\ref{prop:matsuokaduty}.

 \begin{figure*}[ht!]
\subfloat[]{
\includegraphics[width=0.66\columnwidth]{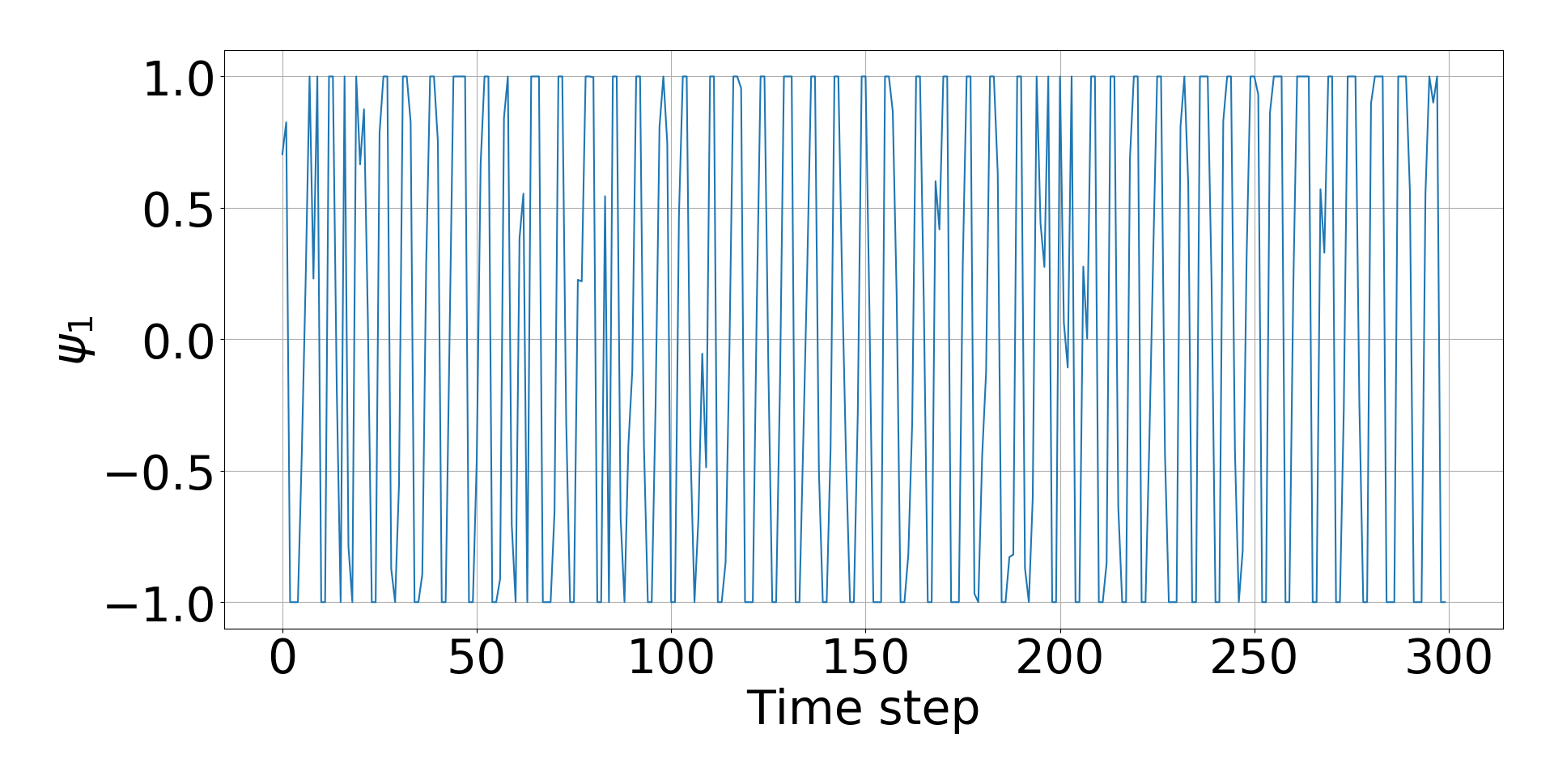}
\label{fig:vanillapsi}}
\subfloat[]{
\includegraphics[width=0.66\columnwidth]{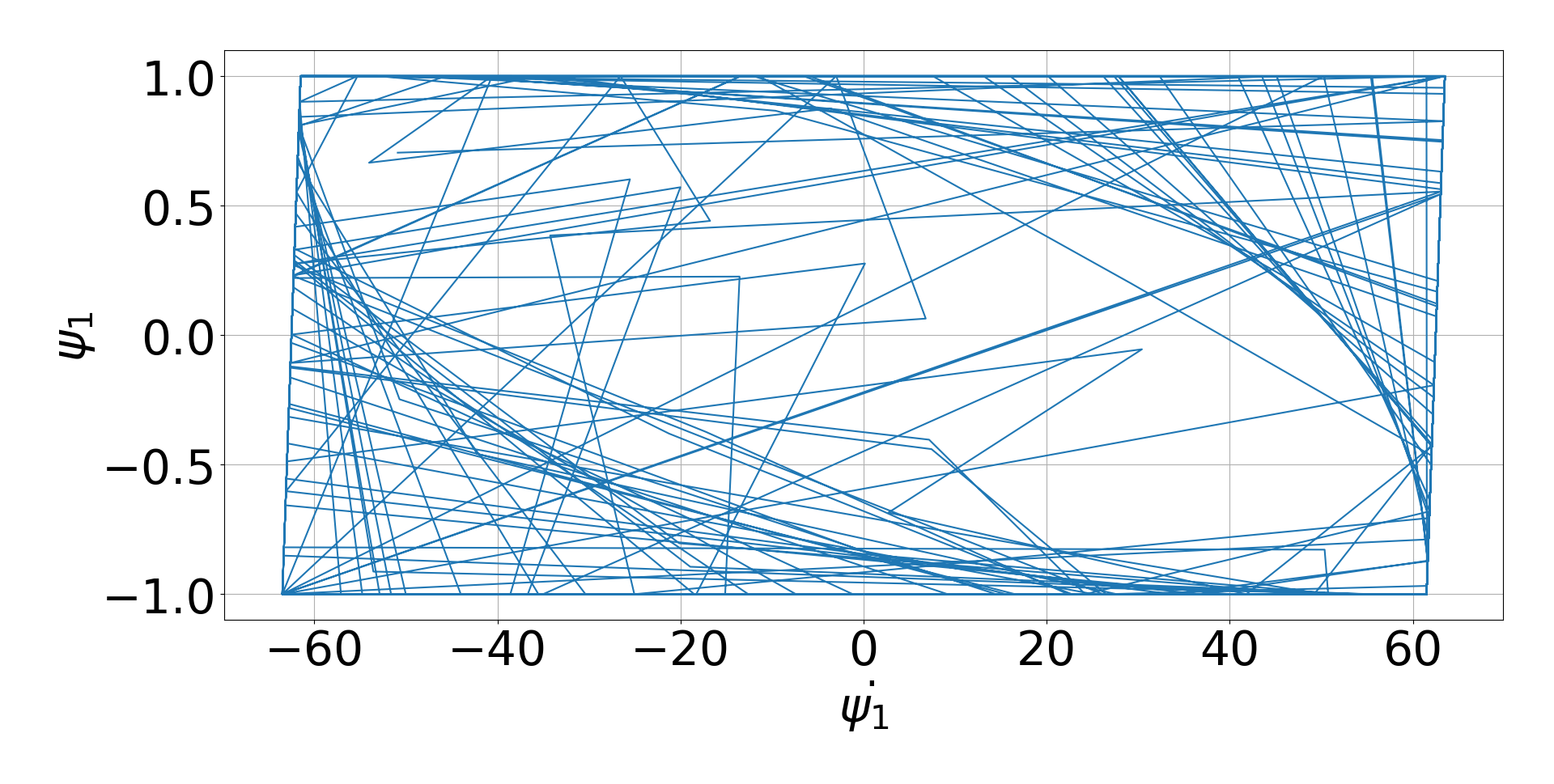}
\label{fig:vanillacycle1}}
\subfloat[]{ 
\includegraphics[width=0.66\columnwidth]{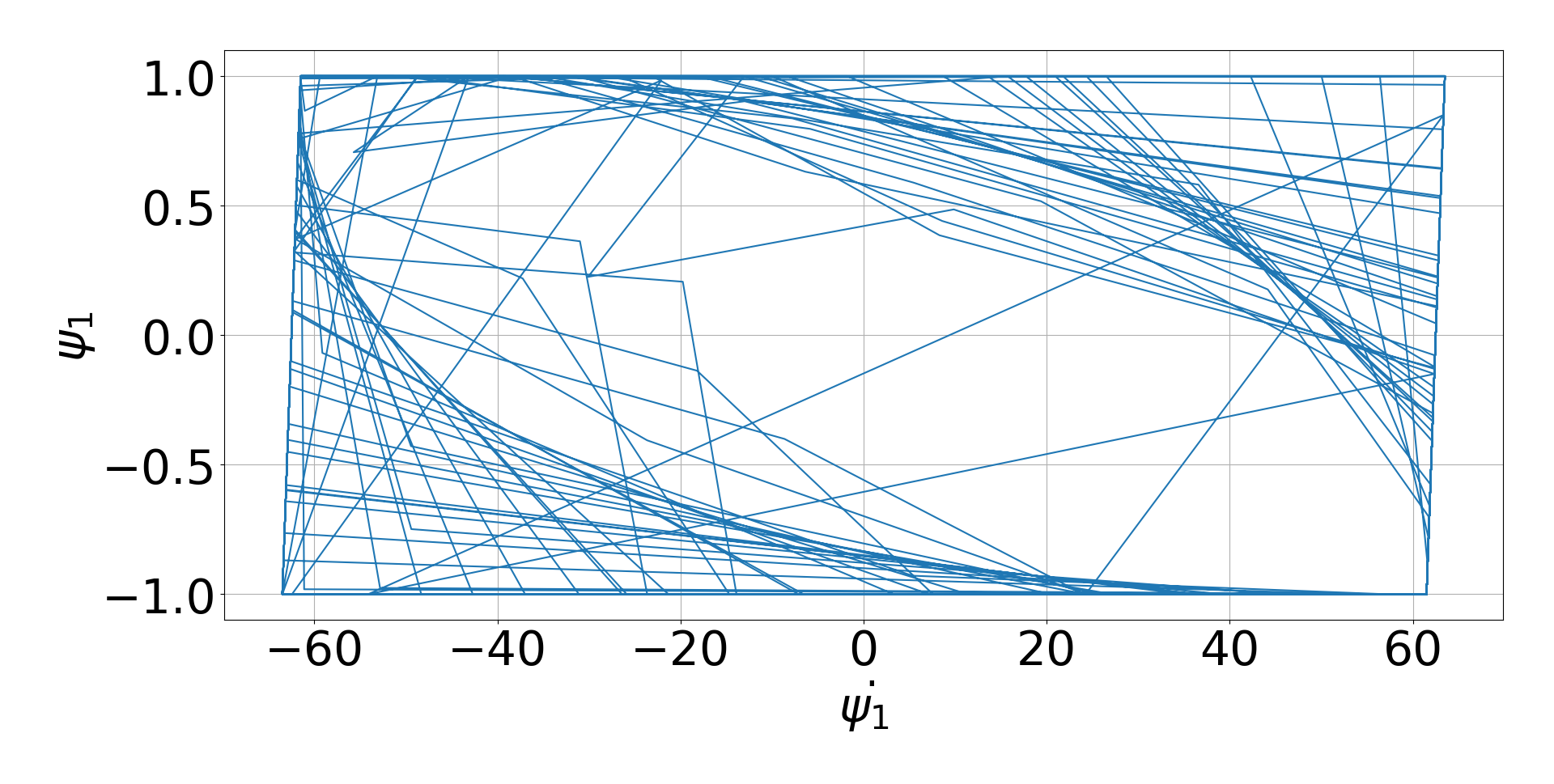}
\label{fig:vanillacycle2}}\\
\vfill
\subfloat[]{ 
\includegraphics[width=0.66\columnwidth]{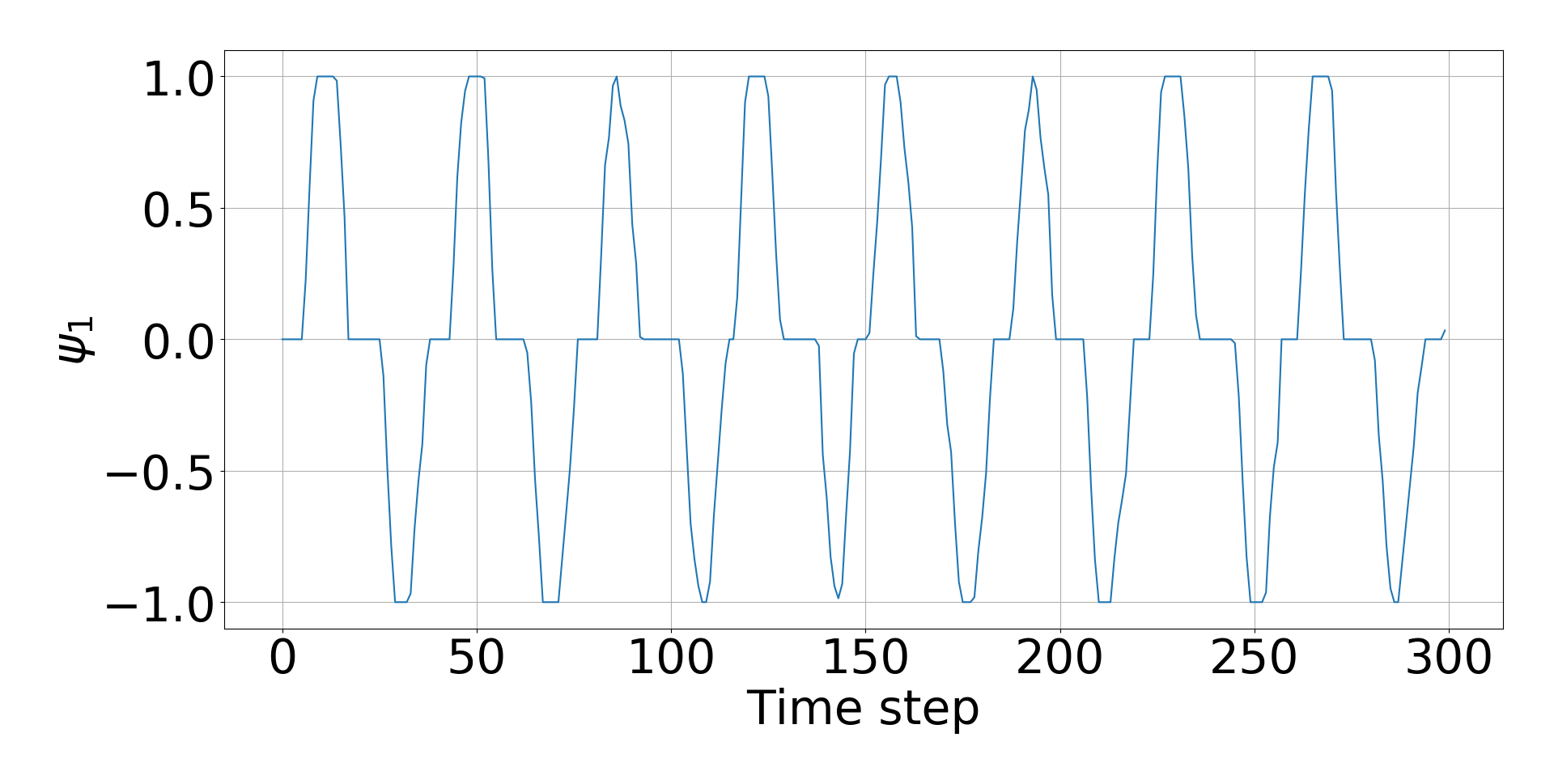}
\label{fig:cpgpsi}}
\subfloat[]{
\includegraphics[width=0.66\columnwidth]{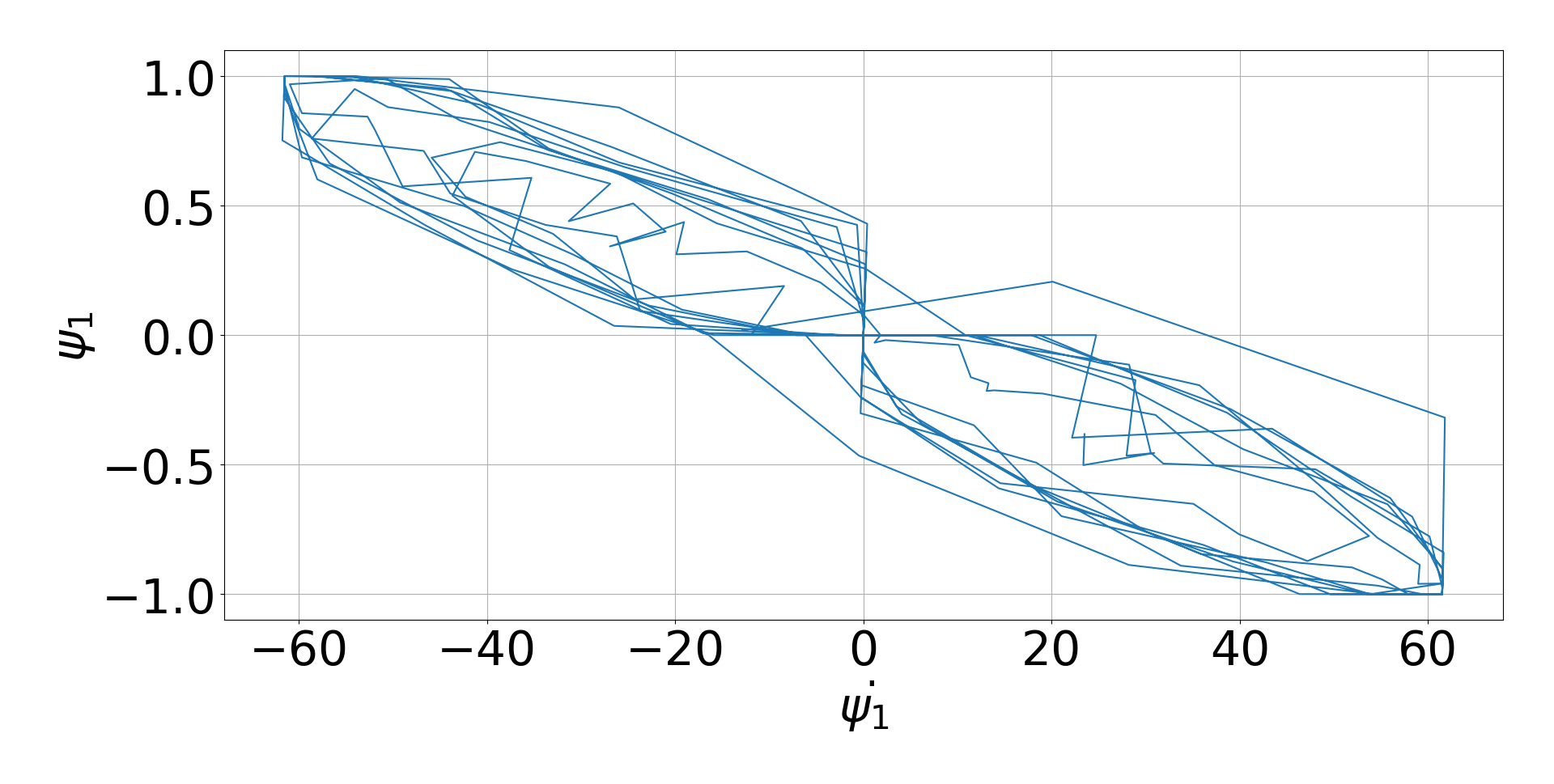}
\label{fig:cpgcycle1}}
\subfloat[]{ 
\includegraphics[width=0.66\columnwidth]{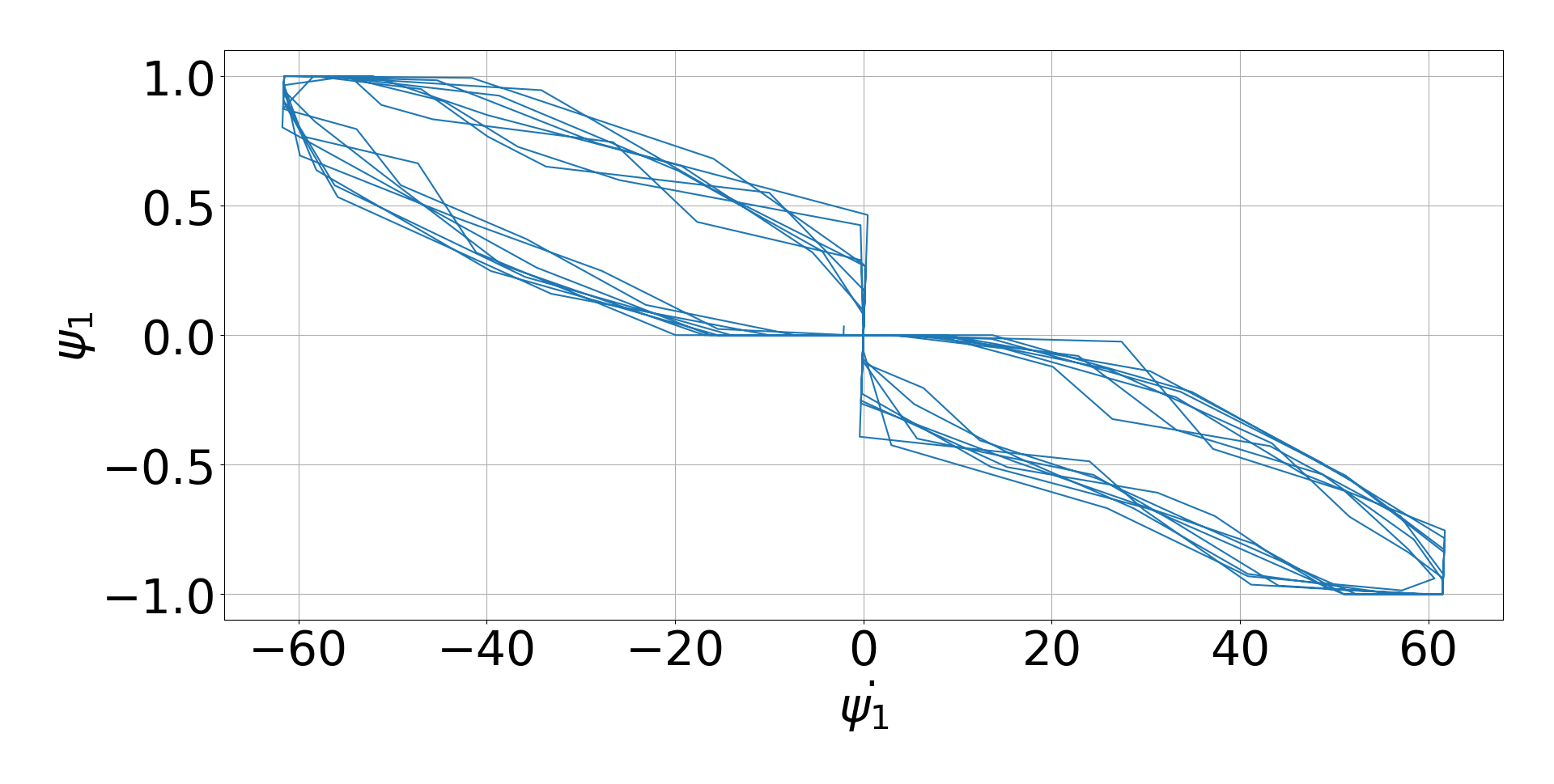}
\label{fig:cpgcycle2}}
\caption{Sample actuation signal $\psi_1$ for the first link generated by (a) vanilla PPO and (d) PPOC-CPG from time step $0$ to time step $300$. Followed by phase plane portraits of $\psi_1$ (b) by vanilla PPO from time step $0$ to $300$, (e) by PPOC-CPG from time step $0$ to $300$, (d) by vanilla PPO from time step $400$ to $700$, (f) by PPOC-CPG from time step $400$ to $700$.}
\label{fig:comparepsi}
\end{figure*}

\subsection{Control signal comparison between PPOC-CPG and vanilla PPO }

First, we compare PPOC-CPG and vanilla PPO in terms of the smoothness of the control input learned in simulation. We train both PPOC-CPG and vanilla PPO in the same environment until convergence. Figure~\ref{fig:comparepsi} shows segments of the control signal $\psi_1$ generated by the vanilla PPO controller and PPOC-CPG controller controlling the simulated soft snake robot in a straight line goal-tracking task. From Fig.~\ref{fig:vanillapsi} it is observed that the signal generated by the vanilla PPO policy oscillates at a relatively higher frequency (about 10Hz on average) with irregular oscillation patterns. Such kind of control signals are not feasible for the actuators in reality. This is because the inflation and deflation of soft air chambers on the snake robot have a certain delay so that the soft pneumatic actuators are not able to track fast oscillating signals. On the other end, the curve in Fig.~\ref{fig:cpgpsi} shows that the agent trained with PPOC-CPG can converge to a stable limit cycle trajectory at a relatively lower but more natural frequency (1.6Hz) for serpentine locomotion. Our approach shows its advantage of being able to generate smoother oscillatory control signals even when the inputs to the CPG system are discontinuous. 
Fig.~\ref{fig:comparepsi} also compares the phase plane portraits recorded at different time stages of the two learning methods. From Fig.~\ref{fig:vanillacycle1} and Fig.~\ref{fig:vanillacycle2}, we observe that the oscillating signal generated by vanilla PPO policy performs irregular oscillation in the first $300$ time steps, and cannot converge to a stable limit cycle when it evolves to time step $700$. While in Fig.~\ref{fig:cpgcycle1} and Fig.~\ref{fig:cpgcycle2}, despite a little deviation from the first $300$ time steps, the outputs of the CPG network eventually converge to a stable limit cycle within $700$ time steps. This experiment shows that the CPG system is capable of stabilizing the oscillation pattern in simple locomotion tasks for the soft snake robot.

\subsection{Comparison of our reward design and a sparse reward function}

\begin{figure}[h!]
    \centering
    \includegraphics[width=0.8\columnwidth]{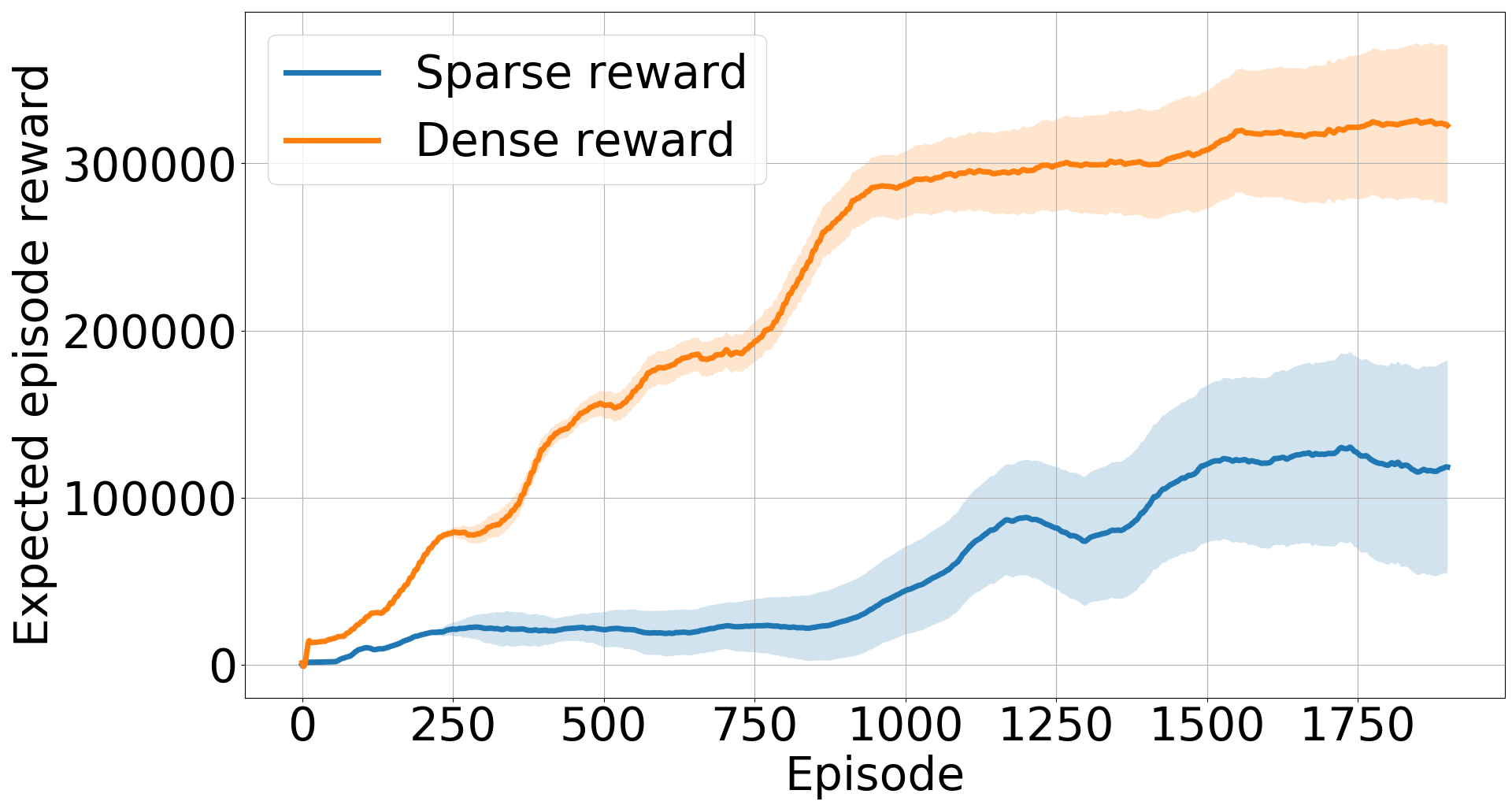}
    \caption{Learning process of FOC-PPOC-CPG with dense reward and sparse reward.}
    \label{fig:comparerewards}
    \vspace{-2ex}
\end{figure} 

We compare the learning process of the revised reward function with our previous one that only rewards the agent for goal reaching events\cite{xliu2020} (for each case we record $5$ learning trials). In average, the agent with dense reward is able to reach and converge to level-$12$, while the agent with sparse reward only converges to level-$8$ (see Table~\ref{tab:curriculum}). The calculated results in Fig.~\ref{fig:comparerewards} show that the system trained with dense reward function outperforms that with a sparse reward design.

In the next section (Section~\ref{sec:focsin2real}), these methods are compared in the real robot to demonstrate the advantage of the proposed PPOC-CPG control.

\begin{figure*}[h!]
    \centering
    \includegraphics[width=0.9\textwidth]{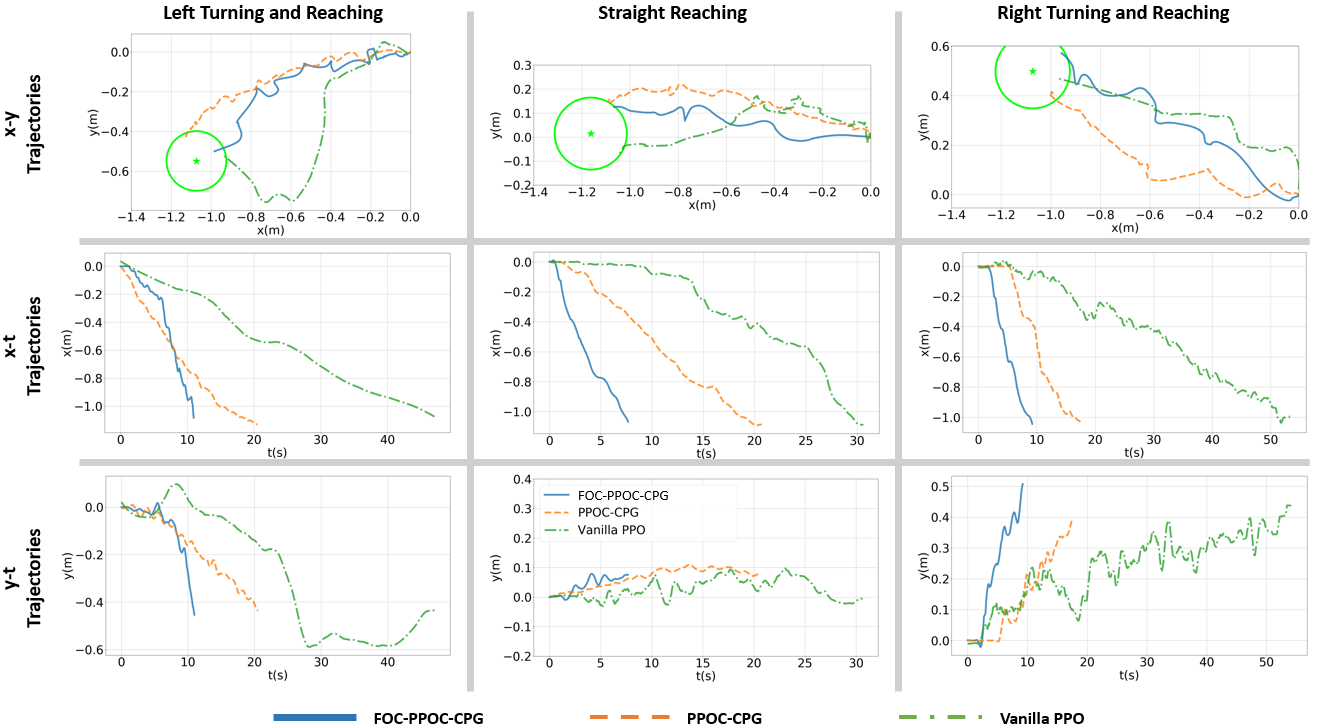}
    \caption{Sample comparison of trajectories generated by Vanilla PPO policy, PPOC-CPG policy, and FOC-PPOC-CPG policy in reality.}
    \label{fig:comparemethods}
\end{figure*}

\begin{table*}[h]
    \centering
    \caption{Performance Comparison of Different Approaches.}
    \label{tab:comparedata}
    \scalebox{0.9}{
    \begin{tabular}{p{4.cm}|p{3.cm}|p{3.cm}|p{3.cm}}

     \hline 
     \textbf{Metrics} & \textbf{Vanilla PPO} & \textbf{PPOC-CPG} & \textbf{FOC-PPOC-CPG} \\  \hline

        Simulated average speed (m/s) & \textbf{0.14} & 0.137& 0.135\\ 

        Simulated success rate & 0.95 & \textbf{0.99}& 0.98\\
        Real average speed (m/s) & 0.027 ($\downarrow 80.7\%$)  & 0.063 ($\downarrow 54\%$)& \textbf{0.121} ($\downarrow 11\%$)\\ 

        Real success rate & 0.5 ($\downarrow 42.1\%$) & 0.82 ($\downarrow 17.1\%$)& \textbf{0.9}($\downarrow 8.1\%$)\\[1ex]
     \hline
    \end{tabular}
}\end{table*}

\begin{figure*}[h!]
\centering
\subfloat[]{
\includegraphics[width=0.7\columnwidth]{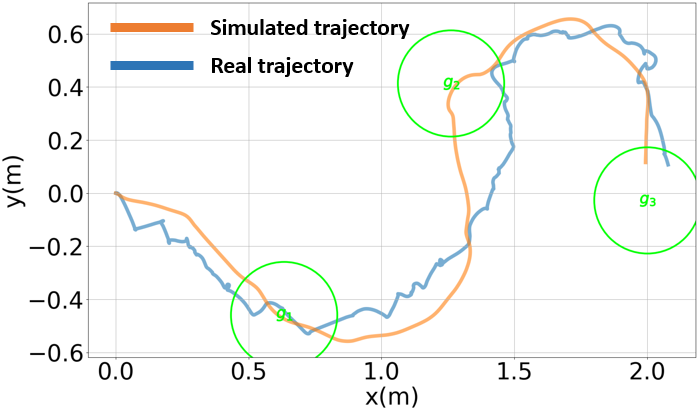}
\label{fig:comparezigzag_sim2real}}
\hspace{0.1\textwidth}
\subfloat[]{
\includegraphics[width=0.4\columnwidth]{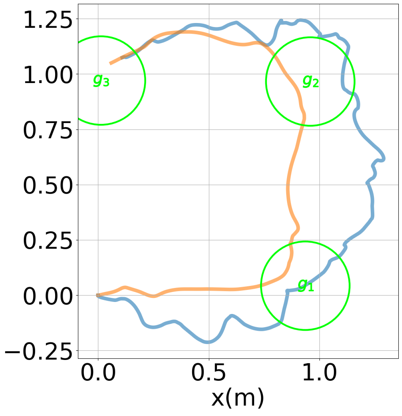}
\label{fig:comparesquare_sim2real}}

\caption{Sample way-point trajectories followed by improved PPOC-CPG controller in simulation and real in (a) zigzag and (b) square.}
\label{fig:comparesimtrajs_sim2real}
\end{figure*}

\subsection{Sim-to-real Performance of FOC-PPOC-CPG}
\label{sec:focsin2real}

\subsubsection{Performance comparison with original PPOC-CPG and Vanilla PPO}
Since FOC-PPOC-CPG is designed for improving the sim-to-real transfer learning performance of the PPOC-CPG method, we first compare the sim-to-real performance of the FOC-PPOC-CPG with the original PPOC-CPG and Vanilla PPO in single goal-reaching tasks. For the real robot tests, all three controllers trained by the simulator are directly applied without further training. We test the controllers by setting goals in three directions (mid, left and right) with fixed angles, distances, and an accuracy radius of $r=0.175$ meters. Each direction takes 10 trials for all three methods in both simulation and reality.

To evaluate the sim-to-real performance, we calculate the average locomotion speed ($v_g$) and the success rate for goal-reaching tasks collected from both simulation and real experiments. According to Section~\ref{sec:FOC}, the contact resistance forces in the simulator are smaller than in the real environment, when applying the RL control policy learned in the simulator directly to the real robot, the performance of the real robot is often worse than the simulated agent. In the rows of real robot evaluations in Table~\ref{tab:comparedata}, we use down-arrows and percentage values to show the extent of performance drop compared to the simulating performance with the same method. 
From Table~\ref{tab:comparedata}, it is observed that although the Vanilla PPO controller learns the best locomotion speed in the simulator at the cost of goal-reaching accuracy, its locomotion pattern cannot fit the real robot well. The real robot experiences a drastic drop in performance on both locomotion speed ($80.7\%$) and success rate ($42.1\%$). For the original PPOC-CPG, though it has achieved an overall better performance than Vanilla PPO, its sim-to-real performance drop is still relatively high, with a $54\%$ of speed drop and $17\%$ of accuracy drop. After adding the free-response oscillation constraint to the CPG system, the new policy reaches almost the same performance as the original PPOC-CPG in the simulator. In Section~\ref{sec:FOC} we have shown that the free-response oscillation tonic input $c>0$ could help maintain the oscillation amplitude of the control signal of FOC-PPOC-CPG during the learning process. It is noticed that the maintained amplitude of the control signals does not improve the locomotion speed and goal-reaching accuracy at the training stage in the simulation. However, when the learned policy of FOC-PPOC-CPG is applied to the real robot without further training, it performs significantly better than the previous two methods in both locomotion speed and success rate. 

Figure~\ref{fig:comparemethods} shows a more intuitive result by comparing sample trajectories of the above three methods in different goal-reaching tasks performed on the real robot. The trajectories show that the robot controlled by Vanilla PPO policy moves much slower than the other two. And it moves in a less symmetric way for the left and right turning tasks. While the original PPOC-CPG and FOC-PPOC-CPG show similar symmetry properties in the trajectory shapes, the difference is that the controller trained with FOC-PPOC-CPG  moves almost twice as fast as that trained with PPOC-CPG. This comparison is presented in the video\footnote{All videos in this paper can be viewed from \url{http://shorturl.at/cgms1}} ``PPO Learning methods comparison.mp4".

Since PPO is an on-policy RL algorithm and has been established for many years, we also use a more up-to-date off-policy RL algorithm -- Twin Delayed Deep Deterministic policy gradient (TD3) \cite{CHEN2022} to replace the role of PPO in our framework, and train it with a shorter learning period (2000 episodes) to verify the generality of our approach. The results and a brief discussion can be viewed in our Supplementary document \footnote{The Supplementary document is available at \url{https://shorturl.at/ntAKM}}.


\begin{figure*}[ht]
\centering

\subfloat[]{
\includegraphics[width=0.66\columnwidth]{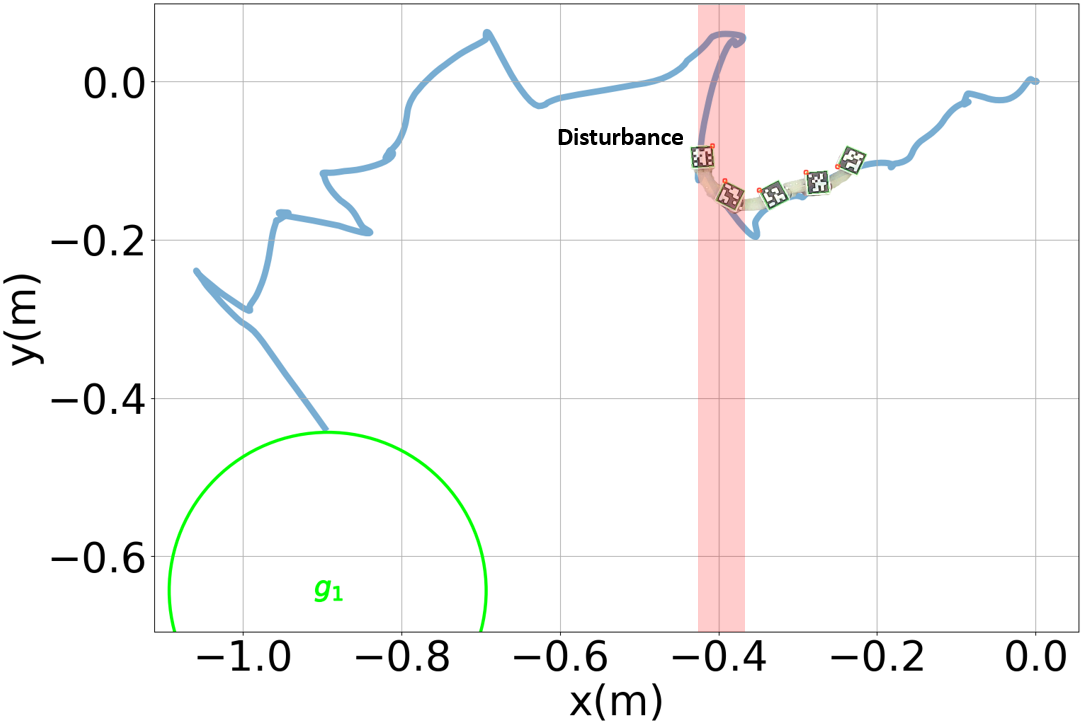}
\label{fig:disturbtraj}}
\hspace{0.1\textwidth}
\subfloat[]{
\includegraphics[width=0.82\columnwidth]{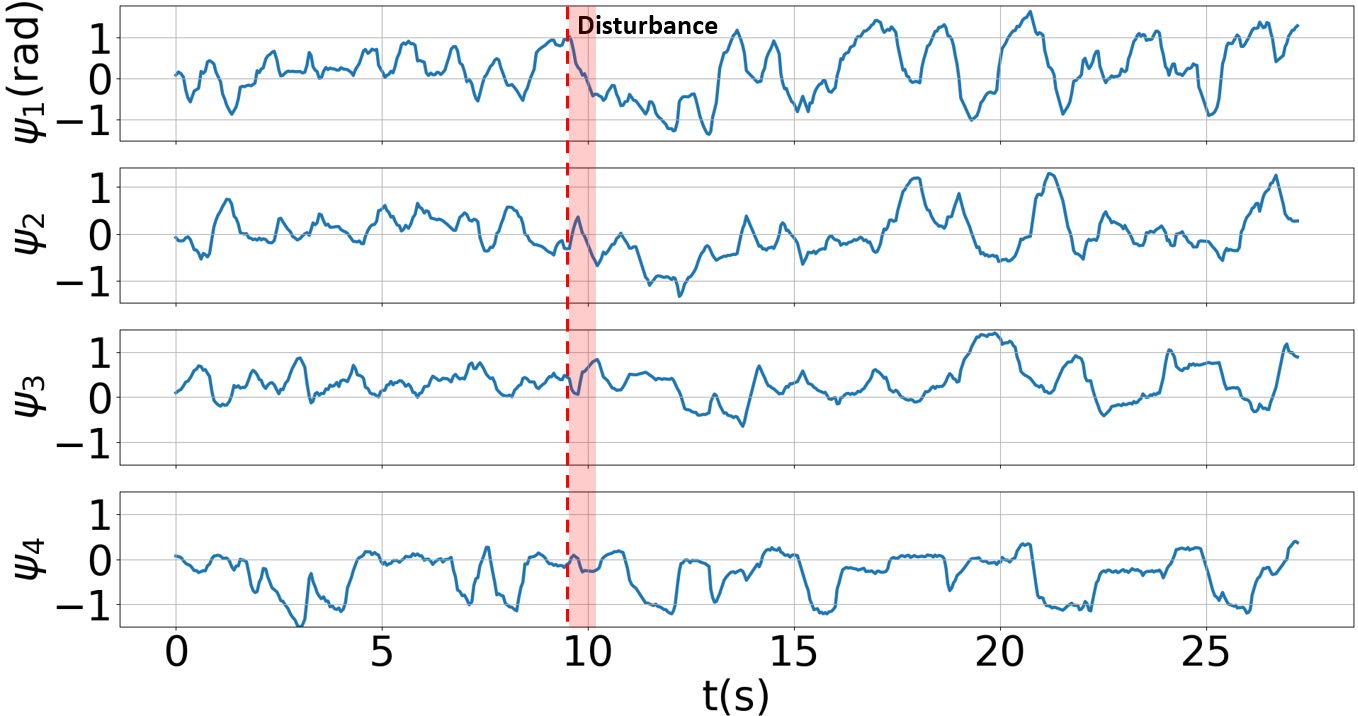}
\label{fig:disturbctrl}}\\

\subfloat[]{
\includegraphics[width=0.92\textwidth]{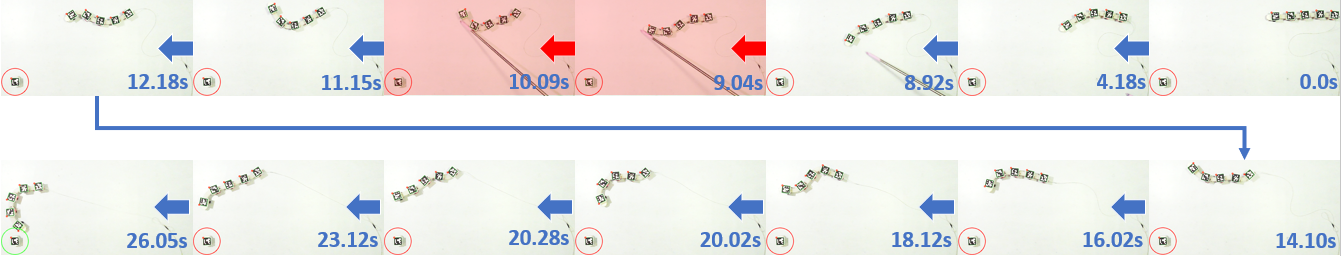}
\label{fig:disturbreal}}

\caption{Disturbance recovery for goal-reaching task followed by FOC-PPOC-CPG controller in real experiments. The presented sub-figures are: (a) x-y plane trajectory, (b) control signals for the actuators, and (c) video snapshot of recorded robot motion.}
\label{fig:comparedisturbtrajs_sim2real}
\end{figure*}

\subsubsection{Performance in reaching unseen goals}

We also investigate the sim-to-real performance of FOC-PPOC-CPG in harder goal-reaching tasks. Figure~\ref{fig:comparesimtrajs_sim2real} compares the head trajectories in Cartesian space for two different setups of way-point goals. The testing trajectories include a square turning trajectory for testing consecutive sharp turning in the same direction (Fig.~\ref{fig:comparesquare_sim2real}), and a zigzag trajectory for testing 
continuous sharp turning in opposite directions (Fig.~\ref{fig:comparezigzag_sim2real}). Both way-point goal series have sharper turning angles than the highest level in the training curriculum in Table~\ref{tab:curriculum}. Video ``Half square trajectory sim2real.mp4" and ``Zigzag trajectory sim2real.mp4" provide the dynamic view of Fig.~\ref{fig:comparezigzag_sim2real} and Fig.~\ref{fig:comparesquare_sim2real} respectively.
From the example videos, it is observed that in both trajectories, the speed drop of the real robot is still around $10\%$, which is not worse than single goal-reaching tasks in Table~\ref{tab:comparedata}. It is noted that in both Fig.~\ref{fig:comparezigzag_sim2real} and Fig.~\ref{fig:comparesquare_sim2real}, it takes the real robot longer distances to make the sharp turning. This is also due to the larger ground resistance forces in reality. 

\subsubsection{Robustness to External disturbance}
We also test the FOC-PPOC-CPG controller's ability to keep tracking the desired target when the robot is disturbed by an external pushing force. Figure~\ref{fig:disturbtraj} and video ``Disturbance recovery.mp4" shows an example trajectory of a disturbed goal-reaching task. It is observed from Fig.~\ref{fig:disturbreal} that the FOC-PPOC-CPG controller reacts accordingly to its situation during the locomotion. When the deviation between the robot's head and the goal is relatively smaller before the disturbance (before $4.1 s$), the robot gently oscillates and adjusts its turning direction gradually towards the goal-direction. At around $9.04s$, when the robot is pushed away from its desired direction, one can observe a clear redirection to the left-hand side of the robot's heading direction. The FOC-PPOC-CPG is able to adjust and make sharp turning to return to the correct direction and still reach the goal without wasting too much time on the recovery. 

\vspace{-2ex}
\section{Conclusion}
\label{sec:conclusion}

This paper develops a bio-inspired controller for learning agile serpentine locomotion with a CPG net mimicking the central nervous system of natural snakes. The contribution of this paper is two-fold: First, we investigate the properties of the Matsuoka oscillator for achieving diverse locomotion skills in a soft snake robot. Second, we construct a FOC-PPOC-CPG net that uses a CPG net to actuate the soft snake robot, and a neural network to efficiently learn a closed-loop near-optimal control policy that utilizes different oscillation patterns in the CPG net. This learning-based control scheme shows promising results in goal-reaching tasks in soft snake robots.

This control scheme can be applicable to a range of bio-mimic motion control for robotic systems and may require different designs of the CPG network given insights from the corresponding biological systems. We have been investigating the generality of the proposed control scheme on different robotic systems and obtained promising early results (see Supplementary document). Our current research focuses on introducing sensory inputs into the CPG system, which enables reactive responses to contact forces with the external environment and generates an obstacle-aided locomotion controller for the soft snake robot. It is also interesting to investigate  distributed control designs that can scale to high-dimensional soft snake robot or other biomimic robotic systems.

\appendices
\numberwithin{equation}{section}

\section{Data}
\label{sec:data}
This section includes the parameter configuration of the Matsuoka CPG network and the hyper parameter setting of domain randomization for the experiment.

\begin{table}[h]
\label{tab:cpgparam}
    \centering
    \caption{Parameter Configuration of the Matsuoka CPG Net Controller for the Soft Snake Robot.}
    \label{tab:config}
    \scalebox{0.9}{
    \begin{tabular}{p{3cm}|p{2cm}|p{2cm}}
     \hline 
     \textbf{Parameters} & \textbf{Symbols} & \textbf{Values} \\  \hline
        Amplitude ratio & $a_{\psi}$ & 2.0935\\ 
        $*$Self-inhibition weight & $b$ & \textbf{10.0355} \\
        $*$Discharge rate & $\tau_r$ & \textbf{0.7696} \\
        $*$Adaptation rate & $\tau_a$ & \textbf{1.7728} \\
        Period ratio & $K_f$ & 1.0\\[1ex] 
        \hline
        Mutual inhibition weights 
         & $a_i$ & 4.6062 \\[1ex]
        \hline
        Coupling weights 
         & $w_{ij}$ & 8.8669 \\
         & $w_{ji}$ & 0.7844 \\ [1ex]
     \hline
    \end{tabular}
}\end{table}

\begin{table}[h]
    \centering
    \caption{Curriculum settings}
    \scalebox{0.9}{
    \begin{tabular}{c|c|c|c}
        \hline
        \textbf{Levels} & \textbf{Distance range ($m$)} & \textbf{Turning angles ($^{\circ}$)} & \textbf{Goal radius ($m$)} \\
        \hline
        1 & $1.2 \sim 1.5$ & $\boldsymbol{-}10 \sim 10$ & 0.5\\
        2 & $1.2 \sim 1.5$ & $\boldsymbol{-}10 \sim 10$ & 0.4\\
        3 & $1.2 \sim 1.5$ & $\boldsymbol{-}15 \sim 15$ & 0.3\\
        4 & $1.2 \sim 1.5$ & $\boldsymbol{-}20 \sim 20$ & 0.25\\
        5 & $1.2 \sim 1.5$ & $\boldsymbol{-}30 \sim 30$ & 0.2\\
        6 & $1.0 \sim 1.5$ & $\boldsymbol{-}40 \sim 40$ & 0.18\\
        7 & $1.0 \sim 1.5$ & $\boldsymbol{-}45 \sim 45$ & 0.15\\
        8 & $1.0 \sim 1.5$ & $\boldsymbol{-}50 \sim 50$ & 0.12\\
        9 & $0.9 \sim 1.5$ & $\boldsymbol{-}60 \sim 60$ & 0.09\\
        10 & $0.9 \sim 1.5$ & $\boldsymbol{-}60 \sim 70$ & 0.06\\
        11 & $0.9 \sim 1.5$ & $\boldsymbol{-}70 \sim 70$ & 0.05\\
        12 & $0.8 \sim 1.5$ & $\boldsymbol{-}80 \sim 80$ & 0.05\\
        \hline
    \end{tabular}}
    \label{tab:curriculum}
\end{table}

\begin{table}[h]
    \centering
    \caption{Domain randomization parameters}
    \scalebox{0.9}{
    \begin{tabular}{c|c|c}
        \hline
        \textbf{Parameter} & \textbf{Low} & \textbf{High} \\
        \hline
        Ground friction coefficient & 0.1 & 1.5 \\
        Wheel friction coefficient& 0.05 & 0.10 \\
        Rigid body mass ($kg$)& 0.035 & 0.075 \\
        Tail mass ($kg$)& 0.065 & 0.085 \\
        Head mass ($kg$) & 0.075 & 0.125 \\
        Max link pressure ($psi$) & 5 & 12\\
        Gravity angle ($rad$) & -0.001 & 0.001\\
        \hline
    \end{tabular}}
    \label{tab:dr}
\end{table}

\section{Foundation}
\subsection{Describing function analysis of the Matsuoka Oscillator}
\label{sec:DF}
According to Fourier theory, we denote the main sinusoidal and constant component in Fourier expansion of the vanilla state $x(t)$ as
\begin{align}
    \label{eq:x}
    x_{\cal F}(t) &= A \cos(\omega t) + d \\ \nonumber
        & = A (\cos(\omega t) + r),
\end{align}
where $r = d/A, r\in R$ is the ratio of bias to the amplitude of the signal. We assume $x_{\cal F}(t)$ only contains cosine term for simplicity. And because this paper only discusses amplitude and bias properties of the signals, such simplification will not affect the following derivations. We use $z_{\cal F}(t) = g(x_{\cal F}(t)) - \epsilon(t) = \max{(x_{\cal F}(t), 0)} - \epsilon(t)$ to represent the main sinusoidal property of $z(t) = g(x(t))=\max{(x(t), 0)}$. In a single period $[-\frac{\pi}{\omega}, \frac{\pi}{\omega}]$,
\begin{align*}
    g(x_{\cal F}(t)) = \begin{cases}
        0 \quad \text{elsewhere}\\
        A(\cos{(\omega t)} + r) \quad t\in[-\frac{\arccos{(-r)}}{\omega}, \frac{\arccos{(-r)}}{\omega}] 
    \end{cases}.
\end{align*}
Using Fourier expansion, the output state $z_{\cal F}(t)$ can also be expressed as:
\begin{align}
\label{eq:z}
    g(x_{\cal F}(t)) \nonumber &= g(A(\cos(\omega t) + r))\\ \nonumber
    &= A g(\cos(\omega t) + r) \\ \nonumber
    &= A(K(r) \cos(\omega t) + L(r)) + \epsilon(t) \\
    &= z_{\cal F}(t) + \epsilon(t)
    \quad (n \geq 1),
\end{align}
such that
\begin{align}
    z_{\cal F}(t) = A(K(r) \cos(\omega t) + L(r)),
\end{align}
where
\begin{align}
\label{eq:kr}
    K(r) = \begin{cases}
    0 \quad (r < -1)\\
    \frac{1}{\pi}(r\sqrt{1-r^2} - \cos ^ {-1}(r)) + 1 \quad(-1 \leq r \leq 1)\\
    1 \quad (r > 1),
    \end{cases}
\end{align}
and 
\begin{align}
\label{eq:lr}
    L(r) = \begin{cases}
    0 \quad (r < -1)\\
    \frac{1}{\pi}(\sqrt{1-r^2} - r\cos ^ {-1}(r)) + r \quad(-1 \leq r \leq 1)\\
    r \quad (r > 1).
    \end{cases}
\end{align}
The derivation of $K(r)$ and $L(r)$ are based on Fourier series analysis (see Appendix \ref{sec:kl}). Both $K(r)$ and $L(r)$ are constrained by $-1\leq r \leq 1$ for $x_{\cal F}(t)$ to be non-negative in the period $[-\frac{\pi}{\omega}, \frac{\pi}{\omega}]$. 

Function $\epsilon(t)$ is the summation of all remaining high frequency terms in the Fourier expansion of $z_{\cal F}(t)$. 

When $t \in [-\frac{\arccos{(-r)}}{\omega}, \frac{\arccos{(-r)}}{\omega}]$, $z_{\cal F}(t) = x_{\cal F}(t)$, we have
\begin{align*}
    &\epsilon(t) = x_{\cal F}(t) - A\{K(r)\cos{(\omega t)} + L(r)\}\\
                &=-\frac{A}{\pi}\{(r
                \sqrt{1-r^2} - \arccos{r})\cos{(\omega t)} + \sqrt{1-r^2} \\  
                & \qquad - r\arccos{r}\}.
\end{align*}

When $t \in [-\frac{\pi}{\omega},-\frac{\arccos{(-r)}}{\omega}]\cup[\frac{\arccos{(-r)}}{\omega}, \frac{\pi}{\omega}]$, $z_{\cal F}(t) = 0$, we have
\begin{align*}
    &\epsilon(t) = 0- A\{K(r)\cos{(\omega t)} + L(r)\}\\
        &= -A\{[\frac{1}{\pi}(r
                \sqrt{1-r^2} - \arccos{r}) + 1]\cos{(\omega t)} \\
                &\qquad - \frac{1}{\pi}(\sqrt{1-r^2} - r\arccos{r})-r\}.
\end{align*}

Then we can numerically calculate the bound of $\epsilon(t)$ for certain $A$ and $\omega$. For instance, if $A=1$ and $\omega=1$, we have
\begin{align*}
    & \epsilon(t)\in[0, 0.2055] \text{ when } \\
    & \qquad t\in[-\frac{\arccos{(-r)}}{\omega}, \frac{\arccos{(-r)}}{\omega}]\\
    & \epsilon(t)\in[-2.0009, 0] \text{ when } \\
    & \qquad t \in [-\frac{\pi}{\omega},-\frac{\arccos{(-r)}}{\omega}]\cup[\frac{\arccos{(-r)}}{\omega}, \frac{\pi}{\omega}].
\end{align*}

\subsection{Calculation of $K(r)$ and $L(r)$}
\label{sec:kl}
Given $x_{\cal F}(t) = A(\cos{(\omega t)} + r)$ as an even function, the general Fourier expansion of $z_{\cal F}(t) = g(x_{\cal F}(t))$ is:
\begin{align}
    z_{\cal F}(t) &= \frac{1}{2} a_0 + \sum_{n=1}^{\infty} a_n \cos(n \omega t) \\ \nonumber
    &= \frac{1}{2} a_0 + a_1 \cos(\omega t) + \epsilon(t).
\end{align}
where 
\begin{align*}
    &a_0 = \frac{1}{\pi} \int_{-\pi}^{\pi} g(A(\cos{(\omega t)} + r)) dt \\
    &a_1 = \frac{1}{\pi} \int_{-\pi}^{\pi} g(A(\cos{(\omega t)} + r))\cos(\omega t) dt.
\end{align*} 

In this case, both the bias $a_0$ and the amplitude $a_1$ become functions of $r$. Combining with \eqref{eq:z}, we use $A K(r)$ to represent $a_1$ and $A L(r)$ to represent $a_0$, which are calculated as follows:
\begin{align*}
    K(r)  = \frac{a_1}{A} &= \frac{\omega}{\pi}\int_{-\pi/\omega}^{\pi/\omega} g((\cos(\omega \tau) + r)) \cos(\omega \tau) d\tau.
\end{align*}
Let $t = \omega \tau$, we have
\begin{align*}
   K(r) &= \frac{1}{\pi}\int_{-\pi}^\pi g((\cos(t) + r)) \cos(t) dt \\
    &= \frac{1}{\pi} \int_{-\cos^{-1}(-r)}^{\cos^{-1}(-r)} (\cos(t) + r) \cos(t) dt \\
    &=  \frac{1}{\pi}(r\sqrt{1-r^2} - \cos ^ {-1}(r)) + 1 \quad(-1 \leq r \leq 1),
\end{align*}
and
\begin{align*}
    L(r)  = \frac{a_0}{A} &= \frac{1}{\pi}\int_{-\pi}^\pi g(\cos(t) + r) dt \\
    &= \frac{1}{\pi} \int_{-\cos^{-1}(-r)}^{\cos^{-1}(-r)} (\cos(t) + r) dt \\
    &=  \frac{1}{\pi}(\sqrt{1-r^2} - r\cos ^ {-1}(r)) + r \quad(-1 \leq r \leq 1).
\end{align*}

\subsection{Derivation of $K_n$}
\label{sec:kn}

Based on \eqref{eq:matsuoka}, we first set $x_i(t) = x_i^e(t) - x_i^f(t)$, $y_i(t) = y_i^e(t) - y_i^f(t)$, $z_i(t) = z_i^e(t) - z_i^f(t)$, $u_i(t) = u_i^e(t) - u_i^f(t)$. Then by taking subtraction between flexor and extensor in \eqref{eq:matsuoka} and neglect phase related coupling terms from other primitive CPGs, we have 
\begin{align*}
     &K_f \tau_r \frac{d}{dt}(x_i^e - x_i^f) = -(x_i^e - x_i^f) - a (z_i^f - z_i^e) \\
     &- b (y_i^e - y_i^f) + (u_i^e - u_i^f)\\ \nonumber
    &K_f \tau_a \frac{d}{dt}(y_i^e - y_i^f) = (z_i^e - z_i^f) - (y_i^e - y_i^f),
\end{align*}
which can be simplified to
\begin{align}
    \label{eq:mergedmatsuoka}
    &K_f \tau_r \frac{d}{dt}x_i = -x_i + a z_i - b y_i + u_i\\ \nonumber
    &K_f \tau_a \frac{d}{dt}y_i = z_i - y_i.
\end{align}

If $x_i^e$ and $x_i^f$ satisfy the \textit{perfect entrainment assumption}, such that the amplitude $A_{x_i^e} = A_{x_i^f} = A_x$, and the bias $r_{x_i^e} = r_{x_i^f} = r_x$, and the phase delay between $x_i^e$ and $x_i^f$ is $\frac{\pi}{\omega}$ (half of the period). Then we have $z_{{\cal F}_i} = K(r_x) x_{{\cal F}_i}$. Similar to the notation in Appendix~\ref{sec:DF}, the marker ${\cal F}_{i}$ indicates the fundamental sinusoidal and constant component in Fourier expansion of the corresponding variable. Let $K_f = 1$, \eqref{eq:mergedmatsuoka} can be further simplified to
\begin{align}
\label{eq:naivematsuoka}
    &\tau_r \frac{d}{dt}x_{{\cal F}_i} + x_{{\cal F}_i} = a K(r_x) x_{{\cal F}_i} - b y_{{\cal F}_i} + u_{{\cal F}_i}\\ \nonumber
    &\tau_a \frac{d}{dt}y_{{\cal F}_i} + y_{{\cal F}_i} = K(r_x) x_{{\cal F}_i}.
\end{align}
Next, an ordinary differential equation can be obtained by merging the two equations in \eqref{eq:naivematsuoka} as, 
\begin{align}
\label{eq:odematsuoka}
    \tau_r \tau_a \frac{d^2}{dt^2} x_{{\cal F}_i} + (\tau_r + \tau_a - \tau_a a K(r_x)) \frac{d}{dt} x_{{\cal F}_i} \\ \nonumber
    + ((b-a) K(r_x) + 1)x_{{\cal F}_i} = \tau_a\frac{d}{dt}u_{{\cal F}_i} + u_{{\cal F}_i}.
\end{align}

When the system is harmonic, the coefficient of the first-order derivative of \eqref{eq:odematsuoka} becomes zero, then 
\begin{align}
\label{eq:appendixkndef}
    K(r_x) = \frac{\tau_r+\tau_a}{\tau_a a} \triangleq K_n.
\end{align}
Coefficient $K_n$ is a special case of $K(r_x)$ in the harmonic Matsuoka oscillator.

\subsection{Amplitude Threshold of Transition from Free Oscillation to Forced Entrainment}
\label{sec:amp}

In order to extract the free-response oscillation component, let $\Tilde{u} = u - 1$, $\Tilde{c} = c+ 1$ then \eqref{eq:matsuoka} is equivalent to
\begin{align}
    \begin{split}
    &K_f \tau_r \Dot{x}_i^e = -x_i^e - a z_i^f - b y_i^e - \sum_{j=1}^N w_{ji}y_j^e + \Tilde{u}_i^e + \Tilde{c},\\ 
    &K_f \tau_a \Dot{y}_i^e = z_i^e - y_i^e,\\ 
    &K_f \tau_r \Dot{x}_i^f = -x_i^f - a z_i^e - b y_i^f - \sum_{j=1}^N w_{ji}y_j^f + \Tilde{u}_i^f + \Tilde{c},\\ 
    &K_f \tau_a \Dot{y}_i^f = z_i^f - y_i^f,
\end{split}\label{eq:purematsuoka}
\end{align}
Since $u_i^q \in [0, 1]$ (for $q\in\{ e,f\}$) and $c\geq 0$, we have $\Tilde{u}_i^q \in [-1,0]$ (for $q\in\{ e,f\}$), and $\Tilde{c} \geq 1$. Now $\Tilde{c}$ becomes the only positive term in the primitive Matsuoka system in \eqref{eq:purematsuoka}. According to Matsuoka's derivation in \cite[(26)]{matsuoka2011analysis}, from \eqref{eq:purematsuoka}, the free-response oscillation amplitude of the Matsuoka oscillator can be written as
\begin{equation}
\label{eq:An}
    A_n = \frac{\Tilde{c}}{K^{-1}(K_n) + (a+b)L(K^{-1}(K_n))}.
\end{equation}

Assume the fundamental harmonic component of the vanilla action signal $\alpha_i$ generated by RL policy has the following form 
\[
    \alpha_{{\cal F}_i} = A \cos{(\omega t)}.
\]
Then substitute $\alpha_{{\cal F}_i}$ into \eqref{eq:decoder}, we have
\begin{align}
     u_{{\cal F}_i}^e \approx \frac{1}{1+e^{-A\cos{(\omega t)}}}.
\end{align}
And
\begin{align}
    \Tilde{u}_{{\cal F}_i}^e \approx \frac{1}{1+e^{-A\cos{(\omega t)}}} - 1.
\end{align}
Because the sigmoid function in $\Tilde{u}_{{\cal F}_i}^e$ is monotonically increasing with $\alpha_{{\cal F}_i}$, the frequency of $\Tilde{u}_{{\cal F}_i}^e$ is the same as the frequency of $\alpha_{{\cal F}_i}$. The amplitude of $\Tilde{u}_{{\cal F}_i}^e$ is
\begin{align}
    A_{\Tilde{u}} &= \frac{\max_t{(\Tilde{u}_{{\cal F}_i}^e(t))} - \min_t{(\Tilde{u}_{{\cal F}_i}^e(t))}}{2} = \frac{1}{2}\frac{e^A - 1}{e^A + 1}.
\end{align}
And the bias of $\Tilde{u}_{{\cal F}_i}^e$ can be calculated as
\begin{align}
    r_{\Tilde{u}} &= \frac{\max_t{(\Tilde{u}_{{\cal F}_i}^e(t))} + \min_t{(\Tilde{u}_{{\cal F}_i}^e(t))}}{2} = -\frac{1}{2}.
\end{align}
It is noted that $\Tilde{u}_{{\cal F}_i}^e$ and $\Tilde{u}_{{\cal F}_i}^f$ are complementary to each other by Definition~\ref{def:complementary}. Thus $\Tilde{u}_{{\cal F}_i}^e$ and $\Tilde{u}_{{\cal F}_i}^f$ share the same amplitude and bias. 

By taking time average of all variables in \eqref{eq:purematsuoka} and ignoring the coupling term from other primitive Matsuoka oscillator nodes, we have the equation of the amplitude of the inner state $x_{{\cal F}_i}^q$ ($q\in\{e,f\}$) as
\begin{align}
\label{eq:Ax}
    A_x[r_x+(a+b)L(r_x)]=\Tilde{c} + r_{\Tilde{u}} = \Tilde{c} - \frac{1}{2}.
\end{align}
Next, since \eqref{eq:purematsuoka} can be reduced to
\begin{align}
\label{eq:purenaivematsuoka}
    &\tau_r \frac{d}{dt}x_{{\cal F}_i} + x_{{\cal F}_i} = a K(r_x) x_{{\cal F}_i} - b y_{{\cal F}_i} + \Tilde{u}_{{\cal F}_i}\\ \nonumber
    &\tau_a \frac{d}{dt}y_{{\cal F}_i} + y_{{\cal F}_i} = K(r_x) x_{{\cal F}_i},
\end{align}
where $\Tilde{u}_{{\cal F}_i} = \Tilde{u}_{{\cal F}_i}^e - \Tilde{u}_{{\cal F}_i}^f$. We derive the describing function from $\Tilde{u}_{{\cal F}_i}(t)$ to $x_{{\cal F}_i}(t)$. Applying the Laplace transform to \eqref{eq:purenaivematsuoka}, we have
\begin{align}
    &G(s, A) = \frac{1}{\tau_r s + 1 - K(r_x)(a - \frac{b}{\tau_a s + 1})} \\ \nonumber
    &= \frac{ \tau_a s + 1}{1 + (\tau_r\tau_a\omega_n^2 - 1)\frac{K(r_x)}{K_n} + \tau_r\tau_a s^2 + (K_n - K(r_x))\tau_a a s}.
\end{align}
More precisely, the frequency transfer function is
\begin{align}
    &G(\omega, A)\\
    &= \frac{ j \tau_a \omega  + 1}{1 + (\tau_r\tau_a\omega_n^2 - 1)\frac{K(r_x)}{K_n} - \tau_r\tau_a \omega^2 + j (K_n - K(r_x))\tau_a a \omega}
\end{align}
where $\omega_n = \frac{1}{\tau_a} \sqrt{\frac{(\tau_a+\tau_r)b}{\tau_r a} - 1}$.
Because the gain from $\Tilde{u}_{{\cal F}_i}(t)$ to $x_{{\cal F}_i}(t)$ is $|G(\omega, A)|$, the amplitude of $x_{{\cal F}_i}(t)$ is given by $|G(\omega, A)|A_u$. Since the amplitude of $x_{{\cal F}_i}(t)$ is twice of $A_x$, and the amplitude of $u_{{\cal F}_i}(t)$ is twice of $A_u$, we have the relation between $A_x$ and $A_u$ expressed as
\begin{align}
    A_x =  |G(\omega, A)| A_{\Tilde{u}} = |G(\omega, A)| A_u.
\end{align}

Given \eqref{eq:odematsuoka}, $(\tau_r + \tau_a - \tau_a a K(r_x))$ is the coefficient of first-order differential variable, also known as damping coefficient. When $K(r_x) = K_n = \frac{\tau_r + \tau_a}{\tau_a a}$, the original oscillation system is harmonic. For the damped oscillation system, the damping coefficient should be positive such that $K(r_x) < K_n$, or equivalently, $\frac{K(r_x)}{K_n} < 1$. In this situation, there will be only forced-response oscillation, and all free-response oscillations diminish due to the positive damping. From \eqref{eq:kr} and \eqref{eq:lr} we know both $K(r)$ and $L(r)$ are monotonic, and therefore $K^{-1}(r)$ and $L^{-1}(r)$ are monotonic as well.
When $K(r_x) < K_n$, 
\begin{align}
    \label{eq:An_forced}
    A_n &= \frac{\Tilde{c}}{K^{-1}(K_n) + (a+b)L(K^{-1}(K_n))} \\ \nonumber
    &< \frac{\Tilde{c}}{r_x+(a+b)L(r_x)},
\end{align}
that is
\begin{equation}
    \label{eq:inequiv_An}
    r_x + (a+b)L(r_x) < \frac{\Tilde{c}}{A_n}.
\end{equation}
From the other end, let $K_x \triangleq K(r_x)$, we have
\begin{align}
\label{eq:Ax_forced}
\footnotesize
&A_x = |G(\omega, A)| A_u\\ \nonumber
&= \frac{A_u\sqrt{\tau_a^2\omega^2 + 1}}{\sqrt{[1+(\tau_r\tau_a\omega_n^2-1)\frac{K_x}{K_n}-\tau_r\tau_a\omega^2]^2 + (K_n-K_x)^2\tau_a^2 a^2 \omega^2}}\\ \nonumber
&\triangleq  \frac{A_u\sqrt{\tau_a^2\omega^2 + 1}}{\sqrt{[1+(\tau_r\tau_a\omega_n^2-1)U-\tau_r\tau_a\omega^2]^2 + K_n^2(1-U)^2\tau_a^2 a^2 \omega^2}}, 
\end{align}
where $U \triangleq \frac{K(r_x)}{K_n}$, and $U\subseteq (0, 1]$.
Next, define a function $Q(U)$ as
\begin{align}
    &Q(U)\triangleq[(\tau_r \tau_a \omega_n^2 - 1)U - (\tau_r \tau_a \omega^2 - 1)]^2 \\ \nonumber
    &\qquad + K_n^2(1-U)^2\tau_a^2 a^2\omega^2.
\end{align}
When $\omega>\omega_n$ and $\tau_r \tau_a \omega_n^2 - 1 > 0$, or $\omega<\omega_n$ and $\tau_r \tau_a \omega_n^2 - 1 < 0$, 
\begin{align}
    Q_{min}(U) = Q(1) = \tau_r^2 \tau_a^2 (\omega^2 - \omega_n^2)^2.
\end{align}
Thus when $U\subseteq (0, 1]$ is satisfied, we have 
\begin{equation}
\label{eq:axlast}
    A_x < A_u\frac{\sqrt{\tau_a^2\omega^2 + 1}}{\tau_r\tau_a|\omega^2 - \omega_n^2|}.
\end{equation}
Combining \eqref{eq:axlast}, \eqref{eq:inequiv_An} and \eqref{eq:Ax}, we have
\begin{align}
    A_u\frac{\sqrt{\tau_a^2\omega^2 + 1}}{\tau_r\tau_a|\omega_n^2 - \omega^2|} \frac{\Tilde{c}}{A_n} > \Tilde{c} - \frac{1}{2} > \Tilde{c} - 1.
\end{align}
Thus we have
\begin{align}
\label{eq:a0accurate}
    A_u > \frac{\Tilde{c} - 1}{\frac{\sqrt{\tau_a^2\omega^2 + 1}}{\tau_r\tau_a|\omega_n^2 - \omega^2|} \frac{\Tilde{c}}{A_n}} = \frac{c}{\frac{\sqrt{\tau_a^2\omega^2 + 1}}{\tau_r\tau_a|\omega_n^2 - \omega^2|} \frac{c+1}{A_n}} \triangleq A_0(c,\omega).
\end{align}
Substitute $A_n$ in the above equation with its approximation in\cite[(30)]{matsuoka2011analysis}, we have
\begin{align}
    A_0(c,\omega) \approx \frac{c}{\frac{\sqrt{\tau_a^2\omega^2 + 1}}{\tau_r\tau_a|\omega_n^2 - \omega^2|} (2K_n-1+\frac{2}{\pi}(a+b)\sin^{-1}(K_n))} 
\end{align}
Since $c \geq 0$, when $\omega$ is fixed, $A_0$ linearly increases with $c$.

\vspace{-2ex}
\section{Theory}

\label{sec:props}
\subsection{Proof of Proposition 1}
\label{sec:prop1}
\begin{IEEEproof} As seen in \eqref{eq:odematsuoka}, when $u_i^e$ and $u_i^f$ of the $i-$th oscillator satisfy constant constraints in Problem 1, the tonic inputs become time-invariant, such that $\frac{d}{dt} u_i(t) = 0$. If the oscillation is harmonic ($K(r_x) = K_n$), then \eqref{eq:odematsuoka} can be rewritten as
    \begin{align}
    \label{eq:bias}
        \tau_r \tau_a \frac{d^2}{dt^2}  x_i + (\tau_r + \tau_a - \tau_a a K_n) \frac{d}{dt} x_i \\ \nonumber
        + ((b-a) K_n + 1) x_i = 2u_i^e - 1,
    \end{align}
Then the above equation can be interpreted as a non-homogeneous spring-damper system with a constant load. By setting
    \[
        \Tilde{x_i} \triangleq x_i - \frac{2 u_i^e - 1}{(b-a) K_n + 1},
    \]
    and substitute $x_i$ with $\Tilde{x_i}$ in \eqref{eq:bias}, we can obtain its homogeneous form as:
    \begin{equation}
        \tau_r \tau_a \frac{d^2}{dt^2}  \Tilde{x}_i + (\tau_r + \tau_a - \tau_a a K_n) \frac{d}{dt} \Tilde{x_i}
        + ((b-a) K_n + 1)\Tilde{x_i} = 0.
    \end{equation}
    Here $\Tilde{x_i}$ is the unbiased variable of $x_i$, and thus the bias of $x_i$ naturally becomes
    \begin{align}
        \mbox{bias}(x_i) = \frac{2 u_i^e - 1}{(b-a) K_n + 1} = \frac{1}{(b-a) K_n + 1}\mbox{bias}(u_i).
    \end{align}

    Since $z_i$ and $x_i$ are entrained (Definition~\ref{def:entrainment}), $z_i = z_i^e - z_i^f = g(x_i^e) - g(x_i^f) = K_n x_i$, we have 
    \begin{align}
        \mbox{bias}(z_i) &= K_n \mbox{bias}(x_i) =  \frac{K_n}{(b-a)K_n + 1}\mbox{bias}(u_i).
    \end{align}
\end{IEEEproof}

\vspace{-5ex}
\subsection{Applicable range of Proposition 1}
\label{sec:prop1limit}
Let $x_i^e< 0$, $x_i^f > 0$, from \eqref{eq:matsuoka}, we have
\begin{align*}
    z_i^e = \max(x_i^e, 0) = 0, z_i^f= \max(x_i^f, 0) = x_i^f.
\end{align*}
Thus
\[  
    z_i = z_i^e - z_i^f = -x_i^f.
\]
Since $u_i^e$ and $u_i^f$ are constants in Proposition~\ref{prop:matsuokabias}, we have
\begin{equation}
\label{eq:xeue}
    x_i^e + a x_i^f = u_i^e + c
\end{equation}
\begin{equation}
\label{eq:xfuf}
    x_i^f + b x_i^f  = u_i^f + c.
\end{equation}
Let $u_i = u_i^e - u_i^f, x_i=x_i^e-x_i^f$, the above two equations can be reduced to
\begin{align}
\label{eq:tempresult}
    u_i = x_i^e + (1+b-a)z_i.
\end{align}
According to Definition~\ref{def:complementary}, $u_i^e+u_i^f=1$, then we have
\[
    u_i^e = \frac{1+u_i}{2}, u_i^f = \frac{1-u_i}{2}.
\]
Substitute $x_i^f$ in \eqref{eq:xeue} with \eqref{eq:xfuf}, and then substitute $u_i^e, u_i^f$ with $u_i$, we have
\begin{align}
    x_i^e = \frac{1+b+a}{2(1+b)} u_i + \frac{1+b-a}{1+b} (\frac{1}{2} + c).
\end{align}
Substitute the above equation of $x_i^e$ to \eqref{eq:tempresult} to obtain
\begin{align*}
    u_i = \frac{1+b+a}{2(1+b)} u_i + \frac{1+b-a}{1+b} + (1+b-a)z_i,
\end{align*}
which can be rearranged to
\begin{align*}
    z_i = \frac{1}{2(1+b)}u_i - \frac{1}{1+b}(\frac{1}{2}+c), \quad (c\geq 0).
\end{align*}

Similarly, for the case when $x_i^e<0, x_i^f>0$, we have
\begin{align*}
    z_i = \frac{1}{2(1+b)}u_i + \frac{1}{1+b}(\frac{1}{2}+c), \quad (c\geq 0).
\end{align*}
Since $z_i$ and $u_i$ are both constants, $\mbox{bias}(z_i)=z_i$ and $\mbox{bias}(u_i)=u_i$. In summary, we have
\begin{align}
\label{eq:prop1compensate}
    \mbox{bias}(z_i) = \begin{cases}
    \frac{1}{2(1+b)}\mbox{bias}(u_i) - \frac{1}{1+b}(\frac{1}{2}+c) \quad (x_i^e< 0, x_i^f > 0)\\
    \frac{1}{2(1+b)}\mbox{bias}(u_i) + \frac{1}{1+b}(\frac{1}{2}+c) \quad (x_i^e> 0, x_i^f < 0).
    \end{cases}
\end{align}

The derivation in this section shows that, when the value of $u_i^e$ and $u_i^f$ causes the Matsuoka system fall into a quadrant such that $x_i^e x_i^f < 0$, the system converges to a set point equilibrium. At this moment the conclusion in Proposition~\ref{prop:matsuokabias} is not applicable to the system. The system should instead follow the relation described in \eqref{eq:prop1compensate}.

The boundary case is at $x_i^e=0, x_i^f>0$ or $x_i^f=0, x_i^e>0$.
For $x_i^e=0, x_i^f>0$, substitute $x_i^e = 0$ to \eqref{eq:xeue} and \eqref{eq:tempresult}, we can obtain the equation
\[
    \mbox{bias}(u_i) = u_i = \frac{2a}{a+b+1} - 1. 
\]
Similarly when $x_i^f>0, x_i^e=0$, we have 
\[
    \mbox{bias}(u_i) = u_i = 1 - \frac{2a}{a+b+1}.
\]

\subsection{Proof of Proposition 2}
\label{sec:prop2}
\begin{IEEEproof} For simplicity we denote $A^q\triangleq A_{x_i^q}$ and $r^q\triangleq r_{x_i^q}$ for $q\in\{e, f\}$. Instead of looking into the relation between $u_i$ and $z_i$, we focus on the bias between the two states. 

According to the \textit{perfect entrainment assumption} \cite{matsuoka2011analysis} and Definition~\ref{def:entrainment}, let $u_i$ be resonant to $z_i$. We define the duty cycle of a wave function as $D(\cdot)$. Let the period of $z_i$ be $T = 2\pi$ (a different value of $T$ would not affect the result of calculation), based on the Fourier expansion, the bias of $u_i$ can be expressed as
\begin{align}
\label{eq: biasu}
    \mbox{bias}(u_i) &= \frac{1}{T}\int_{-T/2}^{T/2} u_i(t) dt \\ \nonumber
    &= \frac{1}{2\pi}\int_{-\pi}^{\pi} u_i(t) dt\\ \nonumber
    &= 2 \frac{1}{2\pi}\int_{-\pi}^{\pi} u_i^e(t) dt - 1 = 2 D(u_i^e) - 1.
\end{align}

Because the bias terms of $x_i$ and $u_i$ are time-invariant, from \eqref{eq:mergedmatsuoka}, we can extract the bias component to form a new equation as follows
\begin{align}
\label{eq:biasraw}
\mbox{bias}(x_i) &= a\cdot \mbox{bias}(z_i) - b\cdot \mbox{bias}(y_i) + \mbox{bias}(u_i)\\ \nonumber
\mbox{bias}(y_i) &= \mbox{bias}(z_i).
\end{align}

Assume $x_i$ can be approximated by its main sinusoidal component and the period of both $x_i$ and $z_i$ is represented by $T$. From \eqref{eq:x} and \eqref{eq:z} we have 
\begin{align*}
    &\mbox{bias}(x_i) = \frac{1}{T}\int_{-T/2}^{T/2} x_i dt = \frac{1}{T}\int_{-T/2}^{T/2} (x_i^e -x_i^f) dt\\
    &\quad = \frac{1}{T}\int_{-T/2}^{T/2} A^e(\cos(\omega t) + r^e) - A^f(\cos(\omega t) + r^f) dt \\
    &\quad = A^e r^e - A^f r^f,\\
    &\mbox{bias}(z_i) = \frac{1}{T}\int_{-T/2}^{T/2} z_i dt = \frac{1}{T}\int_{-T/2}^{T/2} (z_i^e - z_i^f) dt\\
    &\quad = \frac{1}{T}\int_{-T/2}^{T/2} (A^e(K(r^e)cos(\omega t) + L(r^e)) \\
    &\qquad \qquad - A^f(K(r^f)\cos(\omega t) + L(r^f))) dt \\
    &\quad= A^e (L(r^e) - \frac{1}{\pi}) - A^f (L(r_f) - \frac{1}{\pi}) + \frac{1}{\pi} (A^e - A^f).
\end{align*}
Apply Taylor expansion on $L(r)$ (Appendix~\ref{sec:kl}) at $r=0$, we have
\[
    L(r) = \frac{1}{\pi} + \frac{r}{2} + o(r), r\in(-1, 1).
\]
Then we have

\begin{align}
\label{eq:bzbx}
    \mbox{bias}(z_i)&= \frac{1}{2}A^e r^e - \frac{1}{2}A^f r^f +\frac{1}{\pi}(A^e-A^f)\\ \nonumber
    &= \frac{1}{2}\mbox{bias}(x_i) +\frac{1}{\pi}(A^e-A^f).
\end{align}

According to \cite{matsuoka2011analysis}, the amplitude $A^q$ (for $q\in\{e,f\}$) has the form
\[
    A^q = \frac{\mbox{bias}(u^q)+c}{r^q+(a+b)L(r^q)}.
\]
When the system is harmonic, according to \cite[(30)]{matsuoka2011analysis}, we have
\[
    r^e = r^f = K^{-1}(K_n),
\]
such that
\begin{align}
    A^e - A^f &= \frac{\mbox{bias}(u_i^e)-\mbox{bias}(u_i^f)}{ K^{-1}(K_n)+(a+b)L( K^{-1}(K_n))}\\
    &\approx \frac{\mbox{bias}(u_i)}{ 2K_n-1+\frac{2}{\pi}(a+b)\sin^{-1}(K_n)}.
\end{align}

Let $m = \frac{1}{\pi}\frac{1}{2K_n-1+\frac{2}{\pi}(a+b)\sin^{-1}(K_n)}$, \eqref{eq:bzbx} can be rewritten as

\begin{align}
\label{eq:bzbxbu}
    \mbox{bias}(z_i) = \frac{1}{2}\mbox{bias}(x_i) + m\mbox{bias}(u_i).
\end{align}
Substitute $\mbox{bias}(z_i)$ in \eqref{eq:biasraw} with \eqref{eq:bzbxbu}, we can obtain the pure relation between $\mbox{bias}(x_i)$ and $\mbox{bias}(u_i)$ as
\begin{align}
    (1-m(b-a))\mbox{bias}(u_i) = (\frac{1}{2}(b-a)+1) \mbox{bias}(x_i).
\end{align}
In this case, the relation between $\mbox{bias}(z_i)$ and $\mbox{bias}(u_i)$ can be expressed as

\begin{align}
    \mbox{bias}(z_i) &= \frac{1-m(b-a)}{b-a + 2}\mbox{bias}(u_i) + m\mbox{bias}(u_i)\\ \nonumber
    &= \frac{1+2m}{b-a+2}\mbox{bias}(u_i).
\end{align}
\end{IEEEproof}

 
%
\bibliographystyle{IEEEtran}
\bibliography{refs.bib}

\begin{IEEEbiography}[{\includegraphics[width=1in,height=1.25in,clip,keepaspectratio]{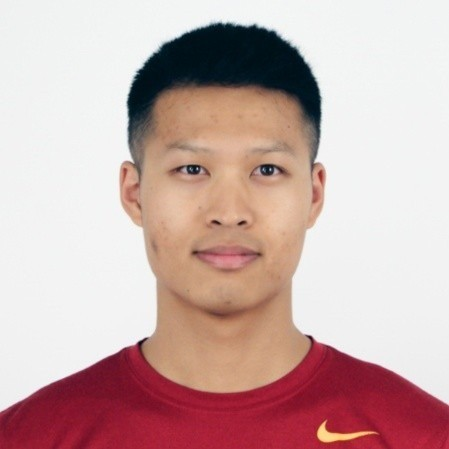}}]{Xuan Liu}
(Student Member, ~IEEE) received the B.S. degree from Beijing University of Posts and Telecommunications, Beijing, China in 2015, and the M.S. degree in computer science from the University of Southern California, Los Angeles, CA, USA. He is currently a Ph.D. student at the Department of Robotics Engineering, Worcester Polytechnic Institute, Worcester, MA, USA.

His research interests include formal methods, stochastic control, reinforcement learning and embodiment control in bio-inspired soft robotics.
\end{IEEEbiography}

\begin{IEEEbiography}
[{\includegraphics[width=1in,height=1.25in,clip,keepaspectratio]{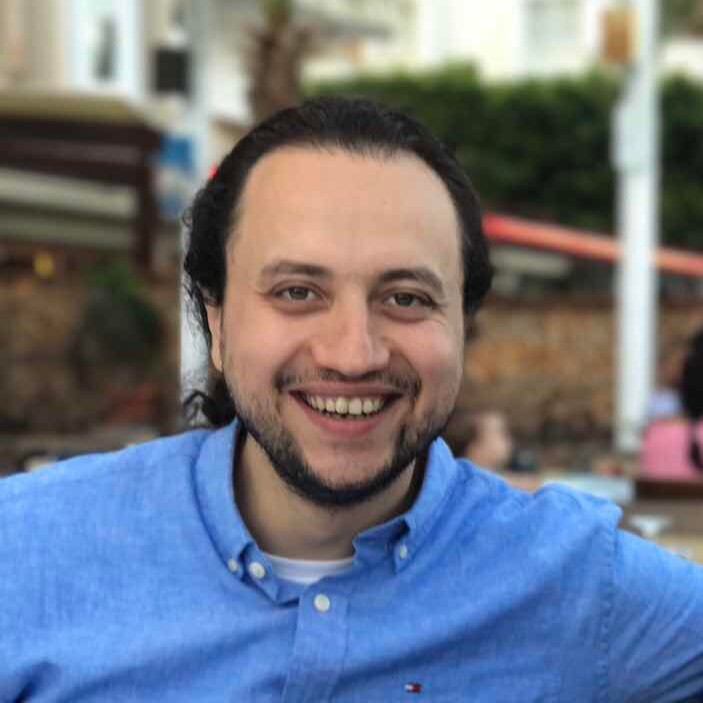}}]
{Cagdas D. Onal}(Member, ~IEEE) is the Dean’s Associate Professor and Arseneault Faculty Fellow in Robotics Engineering at Worcester Polytechnic Institute. He received the B.Sc. and M.Sc. degrees from the Mechatronics Engineering Program, Sabanci University, Turkey, in 2003 and 2005. He received his PhD in Mechanical Engineering from Carnegie Mellon University in 2009. Before joining WPI, he was a Postdoctoral Associate in the Computer Science and Artificial Intelligence Laboratory (CSAIL) at MIT. His research interests include soft robotics, printable robotics, alternative actuation/sensing mechanisms, bio-inspiration, control theory, and wearable robotics. He is the director of WPI Soft Robotics Lab and leads the NSF Future of Robots in the Workplace-Research and Development (FORW-RD) NRT Program at WPI. Prof. Onal is the recipient of an NSF CAREER award on origami-inspired soft robotic systems.

His current research interests include soft robotics, origami-inspired printable robotics, alternative actuation/sensing mechanisms, bio-inspiration, dynamic modeling, and control theory.
\end{IEEEbiography}

\vspace{11pt}
\begin{IEEEbiography}[{\includegraphics[width=1in,height=1.25in,clip,keepaspectratio]{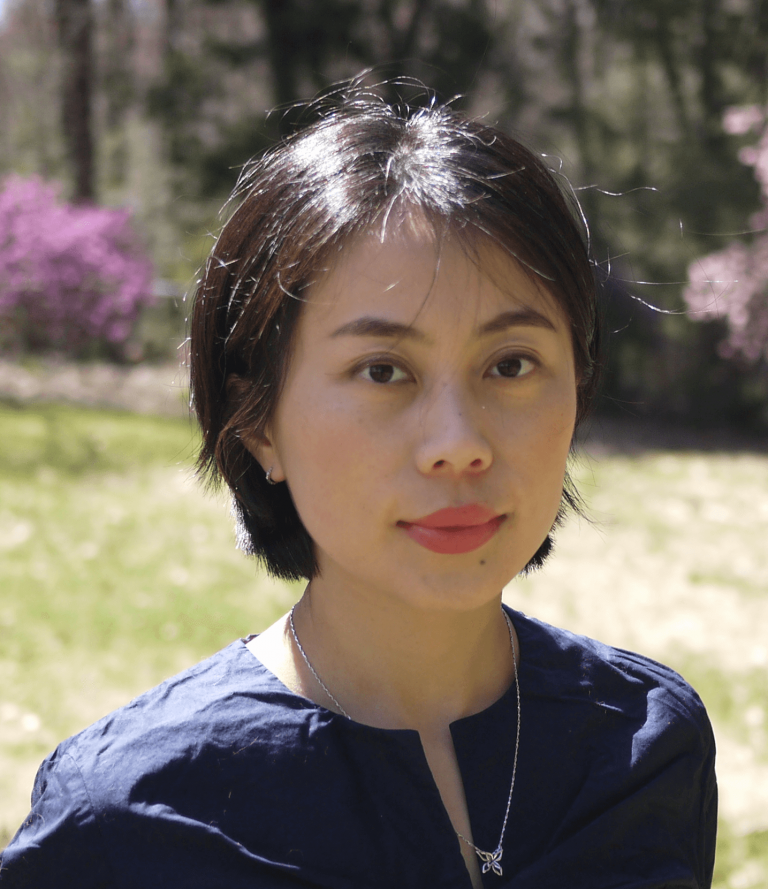}}]{Jie Fu} (Member,~IEEE) received the B.S. and M.S. degrees
from Beijing Institute of Technology, Beijing, China,
in 2007 and 2009, and the Ph.D. degree
in mechanical engineering from the University of
Delaware in 2013. She is currently an Assistant
Professor at the Department of Electrical and Computer Engineering, University of Florida,  Gainesville, FL, USA. 

Her research interests include   stochastic control, reinforcement learning, algorithmic game theory and formal methods.
\end{IEEEbiography}

\vfill

\end{document}